\documentclass[10pt,twocolumn,letterpaper]{article}

\usepackage[pagenumbers]{cvpr} % To force page numbers, e.g. for an arXiv version

\usepackage[dvipsnames]{xcolor}

\definecolor{color3}{RGB}{255, 255, 200}
\definecolor{color2}{RGB}{255, 220, 200}
\definecolor{color1}{RGB}{255, 181, 163}
\newcommand{\cc}[1]{\cellcolor{color#1}}

\usepackage{bm}
\usepackage{multirow}
\usepackage{makecell}
\usepackage{gensymb}
\usepackage{xcolor}
\usepackage{colortbl}
\definecolor{color3}{RGB}{255, 255, 200}
\definecolor{color2}{RGB}{255, 220, 200}
\definecolor{color1}{RGB}{255, 181, 163}
\definecolor{cvprblue}{rgb}{0.21,0.49,0.74}
\usepackage[pagebackref,breaklinks,colorlinks,citecolor=cvprblue]{hyperref}

\def\paperID{9042} % *** Enter the Paper ID here
\def\confName{CVPR}
\def\confYear{2024}

\title{GS-IR: 3D Gaussian Splatting for Inverse Rendering}

\author{Zhihao Liang\textsuperscript{1,*},
Qi Zhang\textsuperscript{2,*},
Ying Feng\textsuperscript{2},
Ying Shan\textsuperscript{2},
Kui Jia\textsuperscript{3,\textdagger}
\\
\textsuperscript{1}South China University of Technology,\quad \textsuperscript{2} Tencent AI Lab,\\
\textsuperscript{3} School of Data Science, The Chinese University of Hong Kong, Shenzhen\\
{\tt\small eezhihaoliang@mail.scut.edu.cn},\quad
{\tt\small nwpuqzhang@gmail.com},\\
{\tt\small vonyfeng@gmail.com},\quad
{\tt\small yingsshan@tencent.com},\quad
{\tt\small kuijia@cuhk.edu.cn}
}
\begin{document}
\maketitle

\let\thefootnote\relax\footnote{* indicates equal contribution.}

\let\thefootnote\relax\footnote{\textsuperscript{\textdagger}Correspondence to Kui Jia $<$kuijia@cuhk.edu.cn$>$.}

%%%%%%%%% ABSTRACT
\begin{abstract}
% imitate from TensoIR, need to modify
We propose GS-IR, a novel inverse rendering approach based on 3D Gaussian Splatting (3DGS) that leverages forward mapping volume rendering to achieve photorealistic novel view synthesis and relighting results. Unlike previous works that use implicit neural representations and volume rendering (\eg NeRF), which suffer from low expressive power and high computational complexity, we extend 3DGS, a top-performance representation for novel view synthesis, to estimate scene geometry, surface material, and environment illumination from multi-view images captured under unknown lighting conditions. There are two main problems when introducing 3DGS to inverse rendering: 1) 3DGS does not support producing plausible normal natively; 2) forward mapping (\eg rasterization and splatting) cannot trace the occlusion like backward mapping (\eg ray tracing). To address these challenges, our GS-IR proposes an efficient optimization scheme incorporating a depth-derivation-based regularization for normal estimation and a baking-based occlusion to model indirect lighting. The flexible and expressive 3DGS representation allows us to achieve fast and compact geometry reconstruction, photorealistic novel view synthesis, and effective physically-based rendering. We demonstrate the superiority of our method over baseline methods through qualitative and quantitative evaluations of various challenging scenes. The source code is available at \url{https://github.com/lzhnb/GS-IR}.
\vspace{-0.6cm}
 
\end{abstract}
\begin{figure}[htbp]
\centering
\includegraphics[width=\linewidth]{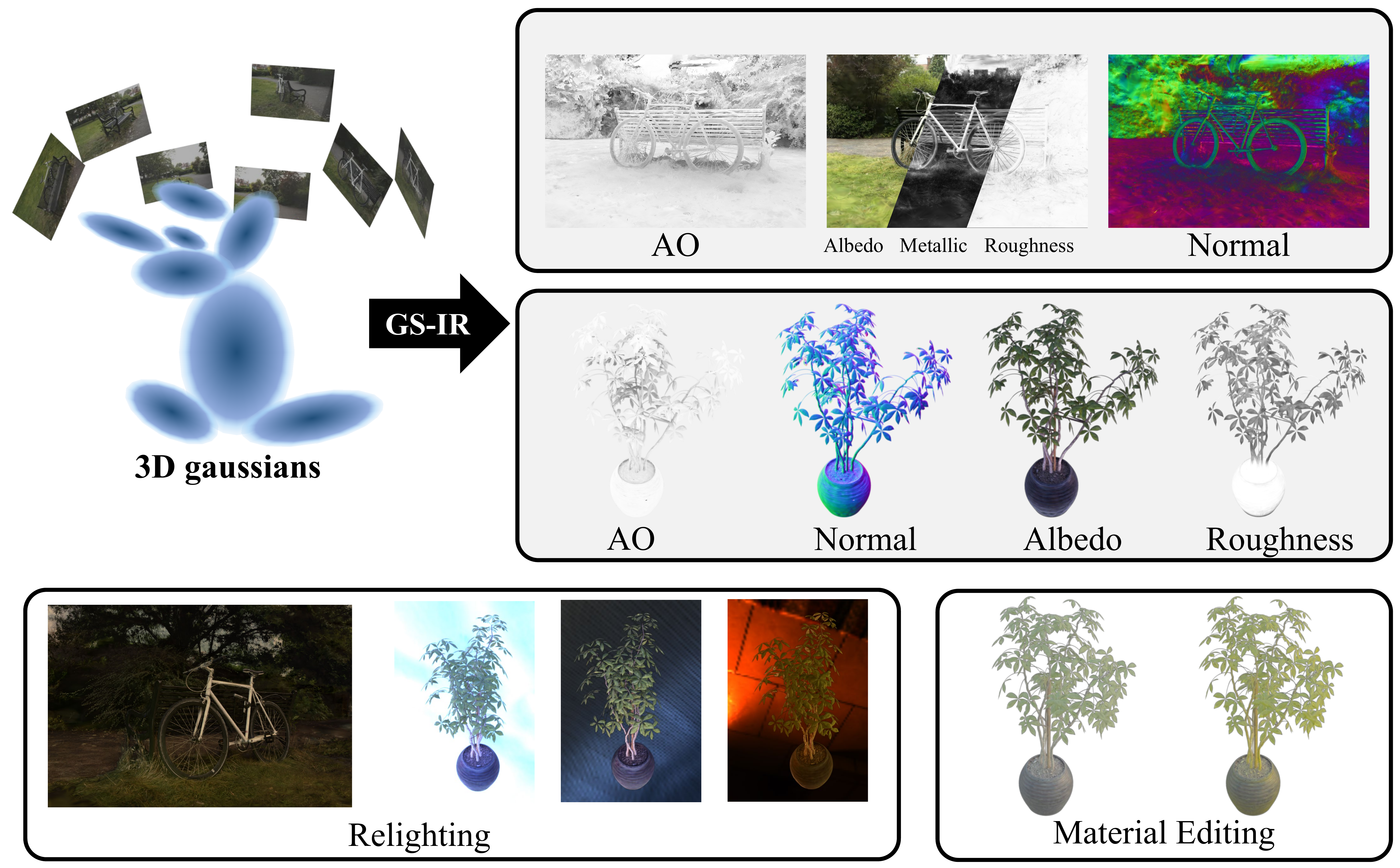}
\caption{Given multi-view captured images of a complex scene, we propose \emph{GS-IR} (3D \textbf{G}aussian \textbf{S}platting for \textbf{I}nverse \textbf{R}endering), which utilizes \textit{3D Gaussian} and forward mapping splatting to recover high-quality physical properties (e.g., normal, material, illumination). This enables us to perform relighting and material editing, resulting in outstanding inverse rendering results. \textbf{Better viewed on screen with zoom in}, especially the \textit{remarkable} material decomposition and normal reconstruction of bicycle axle.}
\label{fig:teaser}
\vspace{-0.6cm}
\end{figure}

%%%%%%%%% BODY TEXT
\section{Introduction}
\label{sec:intro}
Inverse rendering is a long-standing task, seeking to answer the question: ``How can we deduce physical attributes (\eg geometry, material, and lighting) of a 3D scene from multi-view images?". This problem is inherently challenging and ill-posed, particularly when input images are captured in uncontrolled environments with unknown illumination. Recent research \cite{boss2021neural, boss2021nerd, srinivasan2021nerv, zhang2021nerfactor} has sought to address this issue by employing implicit neural representations akin to NeRF \cite{mildenhall2020nerf} that utilizes multi-layer perceptrons (MLPs). However, current methods incorporating MLP face challenges in terms of their low expressive capacity and high computational demands, which significantly limits the effectiveness and efficiency of inverse rendering, especially when it cannot be rendered at interactive rates.

3D Gaussian Splatting (3DGS) \cite{kerbl20233d} has recently emerged as a promising technique to model 3D static scenes and significantly boost the rendering speed to a real-time level. It makes the scene representation more compact and achieves fast and top performance for novel view synthesis. Introducing it to the inverse rendering pipeline is natural and essential, including geometry reconstruction, materials decomposition, and illumination estimation. Unlike ray tracing in NeRF, 3DGS produces a set of 3D Gaussians around sparse points. During the 3DGS optimization, the adaptive control of the Gaussian density may lead to loose geometry, making it difficult to estimate accurate scene's normal. Consequently, it is necessary to introduce a well-designed strategy to regularize 3DGS's normal estimation.

Our goal is to use 3D Gaussians as the scene representation for inverse rendering from multi-view images captured under unknown lighting conditions. However, capturing observations under natural illumination often shows complex effects such as soft shadows and interreflections. TensoIR \cite{jin2023tensoir} leverages the ray tracing of NeRF to directly model occlusion and indirect illumination. In contrast, 3DGS replaces the ray tracing in the NeRF with differentiable forward mapping volume rendering, which directly projects 3D Gaussians onto the 2D plane. This strategy improves the rendering efficiency but makes it difficult to calculate occlusion. Inspired by the  ``\textit{Indirect Lighting Cache}'' used in real-time rendering \cite{akenine2018real}, we attempt to bake the occlusion into volumes for caching.

In this paper, we present a novel 3D Gaussian-based inverse rendering framework called \textit{GS-IR} (3D Gaussian Splatting for Inverse Rendering) that leverages forward mapping splatting to deduce the physical attributes of a complex scene. To the best of our knowledge, our method is the \emph{first} work to introduce the 3DGS technique for inverse rendering, which can simultaneously estimate scene geometry, materials, and illumination from multi-view images. Our GS-IR addresses two main issues when using 3DGS for inverse rendering. Firstly, we develop an intuitive and well-designed regularization to estimate the scene's normal. Secondly, we use a baking-based method embedded in GS-IR to cache occlusions, obtaining an efficient indirect illumination model. As shown in \cref{fig:teaser}, our approach can reconstruct high-fidelity geometry and materials of a complex real scene under unknown natural illumination, enabling state-of-the-art rendering of novel view synthesis and additional applications like relighting.
Our technical contributions are summarized as follows:
\begin{itemize}
\item We present \textit{GS-IR} that models a scene as a set of 3D Gaussians to achieve physically-based rendering and state-of-the-art decomposition results for both objects and scenes;
\item We propose an efficient optimization scheme with regularization to concentrate depth gradient around 3DGS and produce reliable normals for GS-IR;
\item We develop a baking-based method embedded in GS-IR to handle the occlusion in modeling indirect lighting;
\end{itemize}
We demonstrate the superiority of our method to baseline methods qualitatively and quantitatively on various challenging scenes, including the TensoIR synthesis dataset \cite{jin2023tensoir} and Mip-NeRF 360 real dataset \cite{barron2022mip}.

%------------------------------------------------------------------------
\section{Related Works}
\label{sec:formatting}
\noindent\textbf{Neural Representation}
Recently, neural rendering techniques, exemplified by Neural Radiance Field (NeRF) \cite{mildenhall2020nerf}, have achieved impressive success in addressing visual computing problems, giving rise to numerous neural representations \cite{muller2022instant, xu2022point, sun2022direct,chen2022tensorf, yariv2020multiview, huang2024sur2f} tailored for different tasks \cite{chan2022efficient, chan2021pi, li2021neural, cao2023hexplane, fridovich2023k, park2021nerfies, zhang2022arf, liang2023helixsurf}. The vanilla NeRF models a continuous radiance field implicitly in MLPs, which requires massive repeated queries during training and inference. To address the computational inefficiencies, many neural scene representations are proposed with more discretized geometry proxies such as voxel grids \cite{hedman2021baking, sun2022direct, fridovich2022plenoxels}, hash grids \cite{muller2022instant}, tri-planes \cite{chen2022tensorf} or points \cite{xu2022point}. Neural features are stored in a structured manner, allowing for efficient storage and retrieval. The computational cost can thus be significantly reduced by introducing interpolation techniques, however, with an inevitable loss in image quality.
3D Guassians are introduced as an unstructured scene representation to strike a balance between efficiency and quality \cite{kerbl20233d}. With the specially designed tile-based rasterizer for Guassian splats, this method achieves real-time rendering with high quality for novel-view synthesis.
In this work, 3D Gaussian representation is combined with the physical-based rendering (PBR) model for inverse rendering.

\noindent\textbf{Inverse Rendering}
Inverse rendering aims to decompose the image's appearance into the geometry, material, and lighting conditions. Considering the inherent ambiguity between observed images and underlying properties, many methods are proposed with different constrained settings, such as capturing images with fixed lighting and rotating object \cite{dong2014appearance, xia2016recovering}, capturing with moving camera and co-located lighting \cite{bi2020deep, bi2020deep2, luan2021unified, nam2018practical}.
Combined with neural representations, inverse rendering models the scene simulating how the light interacts with the neural volume with various material properties, and estimates the lighting and material parameters during optimization \cite{bi2020neural, srinivasan2021nerv, zhang2021physg, boss2021nerd, boss2021neural, zhang2021nerfactor, zhang2022modeling, hasselgren2022shape, zhang2022iron, jin2023tensoir}.  Neural Reflectance Fields \cite{bi2020neural} assumes a known point light source and represents the scene as a field of volume density, surface normals, and bi-directional reflectance distribution functions(BRDFs) with one bounce direct illumination. NeRV \cite{srinivasan2021nerv} and InvRender \cite{zhang2022modeling} extend to arbitrary known lighting conditions and train an additional MLP to model the light visibility. PhySG \cite{zhang2021physg} assumes full light source visibility without shadow simulation, and represents the lighting and scene BRDFs with spherical Guassians for acceleration. TensoIR \cite{jin2023tensoir} adopts the efficient TensoRF \cite{chen2022tensorf} representation which enables the computation of visibility and indirect lighting by raytracing, while limited to object-level.
For modeling surface geometry using point clouds, Fuzzy Metaballs (FMs) \cite{keselman2022approximate, keselman2023flexible} offers a great way to render depth from 3D Gaussian using Order Independent Transparency (OIT) and approximate intersection. However, it requires silhouettes as input and struggles to handle intricate geometry (\eg Lego and Ficus) let alone complex scenes.
In this work, we propose a 3DGS-based pipeline to recover the geometry, material, and lighting that is available for both objects and unbounded scenes.

%------------------------------------------------------------------------
\section{Preliminary}
\label{sec:preliminary}
In this section, we give the technical backgrounds and math symbols that are necessary for the presentation of our proposed method in subsequent sections.

\noindent\textbf{3D Gaussian Splatting} (3DGS) \cite{kerbl20233d} is an explicit 3D scene representation in the form of point clouds. Each point is represented as a Gaussian function ${g}$ that approximates the shape of a bell curve, which is defined as,
\begin{equation}
\small
{g}(\bm{x}|\bm{\mu}, \bm{\Sigma}) = e^{-\frac{1}{2} \left( \bm{x}-\bm{\mu} \right)^T \bm{\Sigma}^{-1} \left( \bm{x}-\bm{\mu} \right)},
\label{eq:gaussianfunction}
\vspace{-0.1cm}
\end{equation}
where $\bm{\mu} \in \mathbb{R}^3$ is its mean vector, and $\bm{\Sigma} \in \mathbb{R}^{3 \times 3}$ is an anisotropic covariance matrix. The mean vector $\bm{\mu}$ of a 3D Gaussian is parameterized as $\bm{\mu} = (\mu_x, \mu_y, \mu_z)$, and the covariance matrix $\bm{\Sigma}$ is factorized into a scaling matrix $\bm{S}$ and a rotation matrix $\bm{R}$ as $\bm{\Sigma} = \bm{R} \bm{S} \bm{S}^\top \bm{R}^\top$. $\bm{S}$ and $\bm{R}$ refer to a diagonal matrix $\mathrm{diag}(s_x, s_y, s_z)$ and a rotation matrix constructed from a unit quaternion $\bm{q}$. 
Given a viewing transformation with extrinsic matrix $\bm{T}$ and intrinsic matrix $\bm{K}$, the mean vector $\bm{\mu}$ and covariance matrix $\bm{\Sigma}'$ from the 3D point $\bm{x}$ to 2D pixel $\bm{u}$ is defined as,
\begin{equation}
\small
\bm{\mu}'=\bm{K}\bm{T}[\bm{\mu}, 1]^\top, \ \ 
\bm{\Sigma}'=\bm{J}\bm{T}\bm{\Sigma}\bm{T}^\top\bm{J}^\top,
\label{eq:2d_gaussian}
\vspace{-0.1cm}
\end{equation}
where $\bm{J}$ is the Jacobian matrix of the affine approximation of the perspective projection. Besides, each Gaussian represents the view-dependent color $\bm{c}_i$ via a set of coefficients of spherical harmonics (SH), which is then multiplied by opacity $\alpha$ for volume rendering. We finally obtain the color $\hat{\bm{C}}$ at pixel $\bm{u}$ based on \cref{eq:gaussianfunction} and \cref{eq:2d_gaussian}, 
\begin{equation}
\small
\hat{\bm{C}} = \sum_{i \in N}T_i{g}_i(\bm{u}|\bm{\mu}', \bm{\Sigma}')\alpha_i \bm{c}_i, \ \ T_i=\prod^{i-1}_{j=1}(1-{g}_j   (\bm{u}|\bm{\mu}', \bm{\Sigma}')\alpha_j),
\label{eq:volrend}
\vspace{-0.1cm}
\end{equation}
where accumulated transmittance $T_i$ quantifies the probability density of $i$-th Gaussian at pixel $\bm{u}$.

\noindent\textbf{The Rendering Equation} In GS-IR, we leverage the classic rendering equation to formulate the outgoing radiance of a surface point  $\bm{x}$ with normal $\bm{n}$:
\begin{equation}
\small
L_o(\bm{x}, \bm{v}) = \int_\Omega L_i(\bm{x}, \bm{l}) f_r(\bm{l}, \bm{v}) (\bm{l} \cdot \bm{n}) d\bm{l},
\label{eq:rendering}
\end{equation}
$\Omega$ denotes the upper hemisphere centered at $\bm{x}$, $\bm{l}$ and $\bm{v}$ denote incident and view directions respectively. $L_i(\bm{x}, \bm{l})$ denotes the radiance received at $\bm{x}$ from $\bm{l}$. Notably, we follow Cook-Torrance microfacet model \cite{cook1982reflectance, walter2007microfacet} and formulate the bidirectional reflectance distribution function (BRDF) $f_r$ as a function of albedo $\bm{a} \in [0, 1]^3$, metallic $m \in [0, 1]$, and roughness $\rho \in [0, 1]$:
\begin{equation}
\small
f_r(\bm{l}, \bm{v}) =
\underbrace{(1 - m) \frac{\bm{a}}{\pi}}_\text{diffuse} +
\underbrace{
\frac{
    DFG
}{
    4 (\bm{n} \cdot \bm{l}) (\bm{n} \cdot \bm{v})
} 
}_\text{specular},
\label{eq:brdf}
\end{equation}
where microfacet distribution function $D$, Fresnel reflection $F$, and geometric shadowing factor $G$ are related to the surface roughness $\rho$. We use 3D Gaussians to store these material properties in GS-IR.

\begin{figure*}%[htbp]
\centering
\includegraphics[width=0.95\textwidth]{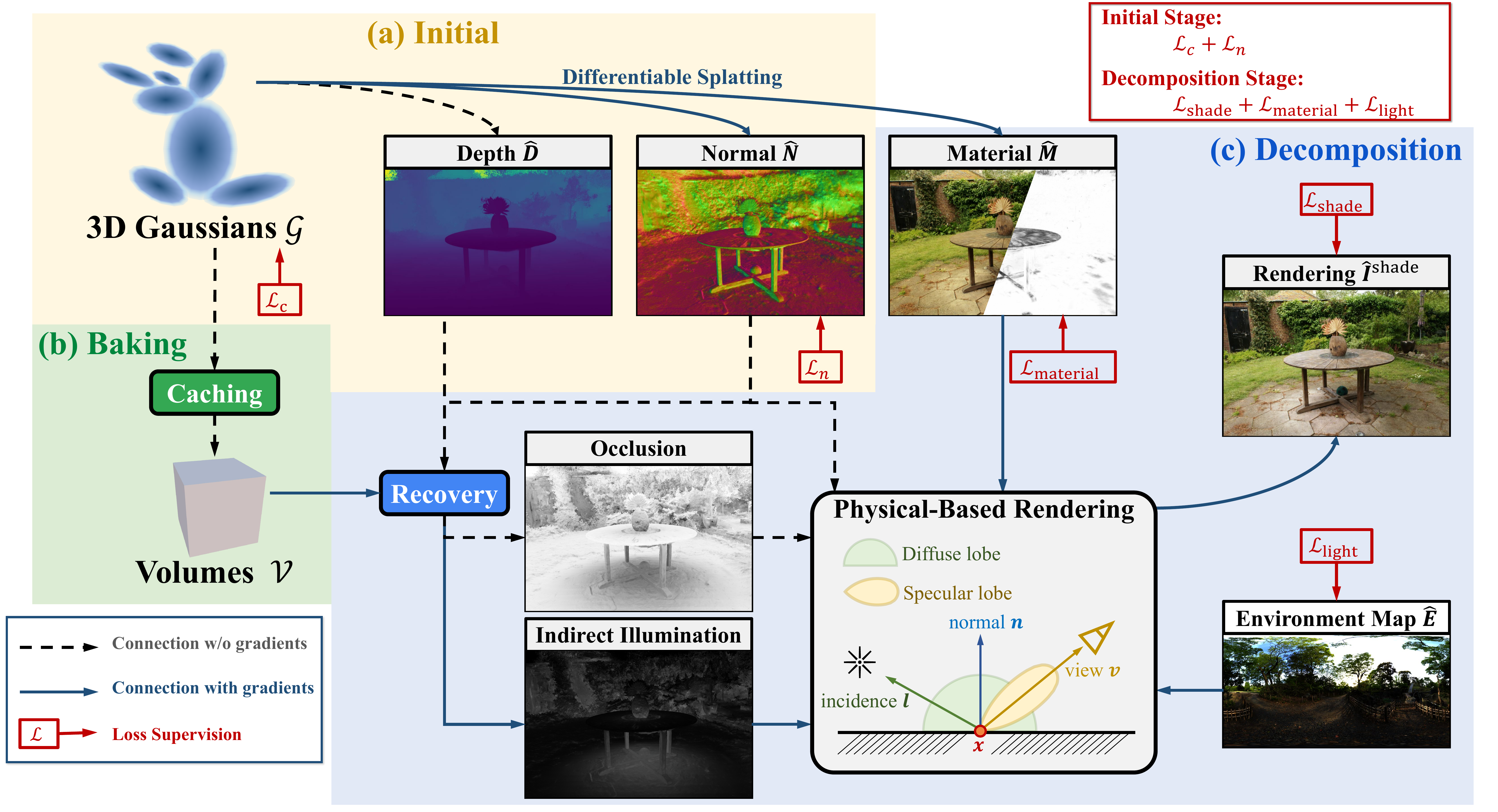}
\caption{\textbf{GS-IR Pipeline}. We propose a novel GS-based inverse rendering framework, called GS-IR, to reconstruct scene geometry, materials, and unknown natural illumination from multi-view captured images. Our GS-IR consists of three well-designed stage strategies using 3D Gaussian and differentiable forward mapping splatting to achieve physical-based rendering. In our approach, the Gaussian stores not only the basic 3DGS information but also the normal and material properties, enhancing its capabilities for inverse rendering tasks. 
% In the initial stage (\cref{subsec:normal}), we 
}
\label{fig:pipeline}
\vspace{-0.4cm}
\end{figure*}

%------------------------------------------------------------------------
\section{Method}
\label{sec:method}
Given a set of calibrated RGB images $\{\bm{I}_m\}^M_{m=1}$ of a target scene captured from multiple views under static, yet unknown illumination, inverse rendering aims to decompose the scene's intrinsic properties, including normal, materials, and illumination. This decomposition facilitates the recovery and subsequent edition of the target scene. Motivated by the remarkable performance in quality and speed of 3DGS \cite{kerbl20233d}, we present a novel framework \textit{GS-IR} consisting of three well-designed stage strategies, as shown in \cref{fig:pipeline}. In the initial stage, we leverage differentiable splatting to optimize 3D Gaussians. Concurrently, we utilize the gradient derived from the rendered depth map to supervise the normal stored in 3D Gaussians (\cf \cref{subsec:normal}). In the second stage, we precompute the occlusion based on the learned geometric information (\ie depth and normal) and store it in an efficient spherical harmonics-based architecture to model indirect illumination (\cf \cref{subsec:indirect}). In the final stage, we combine a differentiable splatting with the physical-based rendering (PBR) pipeline to optimize illumination and material-aware 3D Gaussians (\cf \cref{subsec:decompose}). 
% \cref{fig:pipeline} gives the illustration.

% \subsection{Normal Supervision}
\subsection{Normal Reconstruction}
\label{subsec:normal}
During the initial stage, we optimize 3D Gaussians for geometry reconstruction from observed images, denoted as $\mathcal{G}$. The optimized $\mathcal{G}$ functions as a geometric proxy for surface points and their corresponding normals $\bm{n}$, which are crucial for successful inverse rendering. As highlighted in \cref{sec:intro}, generating reasonable normals within the 3DGS-based framework poses a significant challenge. To address this obstacle, we introduce an intuitive strategy that improves depth $\hat{D}$ and leverages the depth gradient to derive pseudo normals $\hat{\bm{n}}_{\hat{D}} = \nabla_\text{uv}\hat{D}$. These pseudo normals then guide the optimization of normals within the 3D Gaussians.

\begin{figure}%[htbp]
\centering
\includegraphics[width=0.9\linewidth]{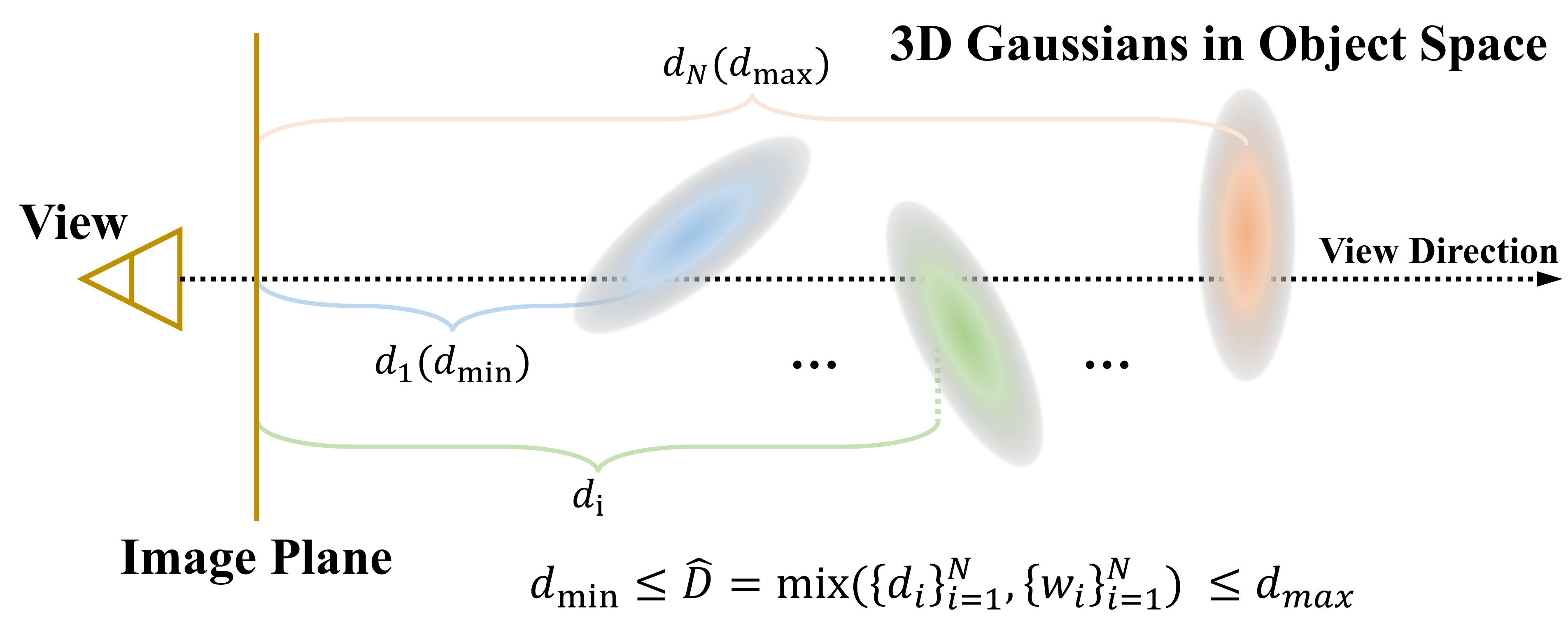}
\caption{\textbf{Depth Illustration}. By considering the depth as a linear interpolation of the distances from 3D Gaussians to the image plane, and ensuring it lies between the minimum and maximum distance, our method could produce accurate depth.}
\label{fig:depth_range}
\vspace{-0.6cm}
\end{figure}

\noindent\textbf{Depth Generation}
Given a pretrained 3D Gaussians $\mathcal{G}$ and a view designated for rendering, the pixel's shading results in that view can be obtained by \cref{eq:volrend}. Consequently, it is reasonable to utilize the same volumetric accumulation to compute the depth $\hat{D}=\sum^N_{i=1} T_i \alpha_i d_i$, where $d_i$ denotes the distance from the corresponding 3D Gaussian to the image plane. However, we observed the \textit{floating} problem during volumetric accumulation, unlike the backward mapping volume rendering used in NeRF. During the 3DGS optimization, the adaptive control of the Gaussian density may result in the depth falling in front of the 3D Gaussians, thereby posing challenges in accurately predicting the depth. Specifically, the backward mapping methods can obtain an accurate depth by considering only peak samples, that is $\hat{D} = d_{i^*}$, where $i^* = \mathop{\arg\max}_{i} T_i \alpha_i$. 
However, for 3DGS, a typical forward mapping method, the peak selection results in \textit{disc aliasing} within 3DGS. To overcome this limitation, we consider that depth $\hat{D}$ must be between the minimum and maximum distance of 3D Gaussians to the image plane, as illustrated in \cref{fig:depth_range}. We then treat the depth as a linear interpolation of the distances from 3D Gaussians to the image plane:
\begin{equation}
\small
\begin{aligned}
\hat{D} = \sum^N_{i=1} \hat{w}_i d_i, \quad \hat{w}_i = \frac{T_i \alpha_i}{\sum^N_{i=1} T_i \alpha_i}.
\end{aligned}
\label{eq:depth_true_accum}
\vspace{-0.1cm}
\end{equation}

\noindent\textbf{Normal Derivation}
% \paragraph{Normal Derivation} 
While the accurate prediction of depth within Gaussians provides better guidance for the normal reconstruction, directly using depth gradient to produce normals has two limitations that still cannot meet the requirement for effective inverse rendering. First, the depth gradient estimation is highly sensitive to noise, making the predicted normal often extremely noisy; Second, the normals derived individually from each view's depth map do not satisfy multi-view consistency. To address these issues, we use Gaussian $\mathcal{G}$ as a proxy for normal estimation instead of directly from the depth gradient. Benefiting from the efficiency of 3DGS, we obtain the depth $\hat{D}$ and normals $\hat{\bm{n}}$ of the observed view after performing a single rendering pass. We then tie these predicted pseudo normal to the underlying depth gradient normal $\hat{\bm{n}}_{\hat{D}}$ using a simple penalty:
\begin{equation}
\small
\mathcal{L}_{n\text{-}p}=\left\|\hat{\bm{n}}-\hat{\bm{n}}_{\hat{D}}\right\|,
\vspace{-0.1cm}
\end{equation}
where $\hat{\bm{n}}=\sum^N_{i=1} T_i \alpha_i \bm{n}_i$, and $\bm{n}_i$ is the normal stored in the 3D Gaussian. Secondly, unlike the MLP-based normal estimation \cite{jin2023tensoir} acting MLP as a low-pass filter, the predicted normal of Gaussian $\mathcal{G}$ is rough, so smoothness regularization should be included. We introduce the TV term $\mathit{TV}_\text{normal}$ to smooth the predicted normal $\hat{\bm{n}}$. For more details, please refer to the \emph{supplement}.

In optimization of the first stage, we optimize 3D Gaussians $\mathcal{G}$ (storing SH coefficients for view-dependent color $\bm{c}$, opacity $\alpha$, and normal $\hat{\bm{n}}$) by using the color reconstruction loss $\mathcal{L}_{c}$, which is the same as 3DGS \cite{kerbl20233d}, and the proposed normal loss $\mathcal{L}_n$,
\begin{equation}
\small
\mathcal{L}_n=\mathcal{L}_{n\text{-}p} + \lambda_{n\text{-}TV}\mathit{TV}_\text{normal}.
\label{eq:normal_loss}
\vspace{-0.1cm}
\end{equation}

\begin{figure}%[htbp]
\centering
\includegraphics[width=0.9\linewidth]{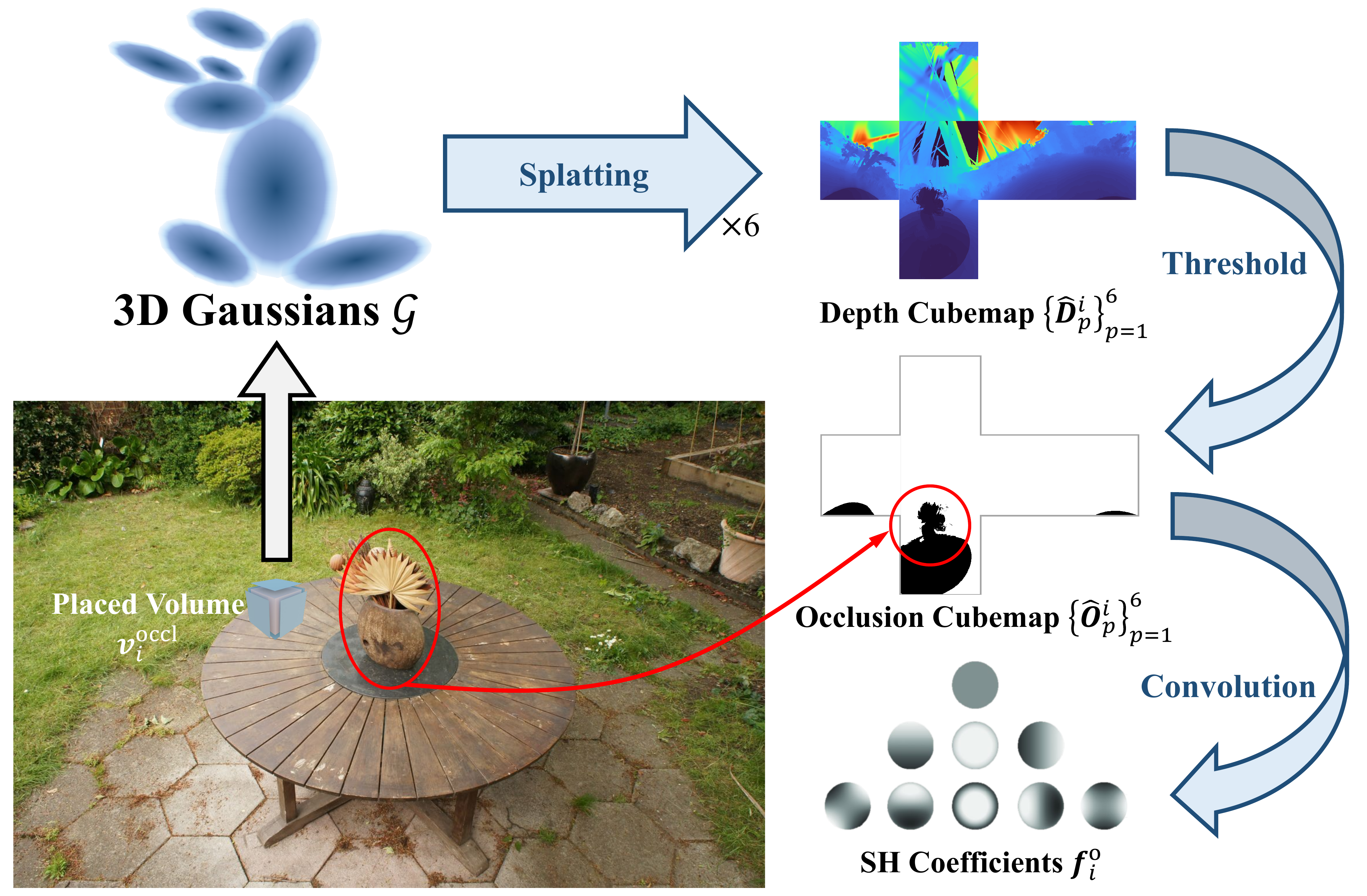}
\caption{\textbf{Baking}. We employ the spherical harmonics (SH) architecture to bake occlusion volumes for modeling indirect illumination. 
For each grid of occlusion volumes, we initially use 3D Gaussians to compute the depth cubemap by performing six forward mapping splatting passes. Next, we convert the depth cubemap into a binary occlusion cubemap based on a distance threshold. Finally, the occlusion cubemap is baked as SH coefficients, enabling efficient interpolation of the occlusion cubemap at any point within the scene.}
\label{fig:baking}
\vspace{-0.6cm}
\end{figure}

\subsection{Indirect Illumination Modeling}
\label{subsec:indirect}
Drawing inspiration from the successful implementation of precomputation techniques in the video game industry (\eg Irradiance Volume in Blender \cite{blender}, Lightmass Volume in Unreal \cite{UE}, and Light Probes in Unity \cite{Unity}), we introduce spherical harmonics (SH) architectures to store occlusion information and model indirect illumination.

Given the optimized 3D Gaussians $\mathcal{G}$ from one stage (\cf \cref{subsec:normal}), we freeze $\mathcal{G}$ and regularly place occlusion volumes $\mathcal{V}^\text{occl}$ in the 3D space. For each volume $\bm{v}^\text{occl}_i \subset \mathcal{V}^\text{occl}$, we then cache the occlusion in the form of SH coefficients $\bm{f}^\text{o}_i$.
Consequently, the formula of the occlusion $O(\cdot)$ of $\bm{v}^\text{occl}_i$ with respect to the direction $(\theta, \phi)$ is expressed as:
\begin{equation}
\small
O(\theta, \phi) = \sum^\mathit{deg}_{l=0} \sum^{l}_{m=-l} \bm{f}^\text{o}_{i(lm)} Y_{lm}(\theta, \phi),
\label{eq:occlusion}
\vspace{-0.1cm}
\end{equation}
where $\mathit{deg}$ denotes the degree of SH, and $\{Y_{lm}(\cdot)\}$ is a set of real basis of SH.

As discussed in \cref{subsec:normal}, the 3DGS technique employs a forward mapping approach that projects 3D points to the 2D plane, in contrast to the backward mapping volume rendering utilized in NeRF, which means it cannot use ray marching to calculate occlusions. 
To precompute the SH coefficients $\bm{f}^\text{o}_i$ of occlusion volume $\bm{v}^\text{occl}_i$, we obtain the depth cubemap $\{\hat{\bm{D}}^i_p\}^6_{p=1}$ by performing six times rendering passes, once for each face of the cubemap. We then convert it into a binary occlusion cubemap $\{\hat{\bm{O}}^i_p\}^6_{p=1}$ based on a manually set distance threshold. 

Finally, we convolve the the occlusion cubemap $\{\hat{\bm{O}}^i_p\}^6_{p=1}$ using SH bases and get the SH coefficients $\bm{f}^\text{o}_i$:
\begin{equation}
\footnotesize
\begin{aligned}
\bm{f}^\text{o}_{i(lm)} =& \int_{S^2} \hat{\bm{O}}(\bm{\omega}) Y_{lm}(\bm{\omega}) d \bm{\omega} \\
=& \int^{2\pi}_0\int^{\pi}_0 \sin\theta\ \hat{\bm{O}}(\theta, \phi) Y_{lm}(\theta, \phi) d\theta d\phi,
\end{aligned}
\label{eq:sh_convolution}
\vspace{-0.1cm}
\end{equation}
where $S^2$ denotes the unit sphere, and $\hat{\bm{O}}(\theta, \phi)$ denotes the occlusion query from the occlusion cubemap $\{\hat{\bm{O}}^i_p\}^6_{p=1}$. Note that we numerically calculate the convolution in \cref{eq:sh_convolution} in parallel.
The caching process is shown in \cref{fig:baking}.

To handle the indirect illumination in occlusion regions, we also maintain illumination volumes $\mathcal{V}^\text{illu}$ to cache the indirect illumination. Similar to the caching process of occlusion volumes, the caching target of illumination volumes changes from the occlusion cubemap $\{\hat{\bm{O}}^i_p\}^6_{p=1}$ to the captured environment cubemap $\{\hat{\bm{I}}^i_p\}^6_{p=1}$. These cubemaps can be obtained simultaneously by conducting six rendering passes. For more details, please refer to the \textit{supplement}.

\subsection{Intrinsic Decomposition}
\label{subsec:decompose}
In the final stage, we employ differentiable splatting in conjunction with a PBR pipeline to accomplish the intrinsic decomposition. According to \cref{eq:brdf}, the rendering equation \cref{eq:rendering} is rewritten as diffuse $L_\text{d}$ and specular $L_\text{s}$ components:
\begin{equation}
\scriptsize	
\begin{aligned}
L_o(\bm{x}, \bm{v}) =& \int_\Omega
\left[(1 - m) \frac{\bm{a}}{\pi} +
\frac{
    DFG
}{
    4 (\bm{n} \cdot \bm{l}) (\bm{n} \cdot \bm{v})
}\right]
L_i(\bm{x}, \bm{l}) (\bm{l} \cdot \bm{n}) d\bm{l} \\
% =& L_\text{diffuse} + L_\text{specular} \\
L_\text{d} =& (1 - m) \frac{\bm{a}}{\pi} \int_\Omega L_i(\bm{x}, \bm{l}) (\bm{l} \cdot \bm{n}) d\bm{l} \\
L_\text{s} =& \int_\Omega
\frac{
    DFG
}{
    4 (\bm{n} \cdot \bm{l}) (\bm{n} \cdot \bm{v})
}
L_i(\bm{x}, \bm{l}) (\bm{l} \cdot \bm{n}) d\bm{l}.
\end{aligned}
\label{eq:shading}
\end{equation}
In GS-IR, we adopt an image-based lighting (IBL) model and split-sum approximation \cite{karis2013real} to tackle the intractable integral.
To calculate the diffuse component $L_\text{d}$, the illumination $I_\text{d}$ is defined as:
\begin{equation}
\footnotesize
\begin{aligned}
I_\text{d}(\bm{x}) =& \int_\Omega L_i(\bm{x}, \bm{l}) (\bm{l} \cdot \bm{n}) d\bm{l} \\
=& \int_{\Omega_\text{Vis}} L^\text{\tiny dir}_i(\bm{x}, \bm{l}) (\bm{l} \cdot \bm{n}) d\bm{l} +
\int_{\Omega_\text{Occl}} L^\text{\tiny indir}_i(\bm{x}, \bm{l}) (\bm{l} \cdot \bm{n}) d\bm{l} \\
\approx& \left(1 - \text{O}(\bm{x})\right) I^\text{\tiny dir}_\text{d}(\bm{x}) + \text{O}(\bm{x}) I^\text{\tiny indir}_\text{d}(\bm{x}),
\end{aligned}
\label{eq:diffuse}
\end{equation}
where the first component indicates direct illumination and the second is indirect illumination. Notably, our baking-based indirect illumination model enables us to calculate the occlusion and illumination online. This means that our GS-IR achieves intrinsic decomposition while maintaining real-time rendering performance. 
For the specular component  $L_\text{s}$, we follow split-sum approximation and treat the integral as two separate integrals:

\begin{figure*}%[htbp]
\centering
\includegraphics[width=\textwidth]{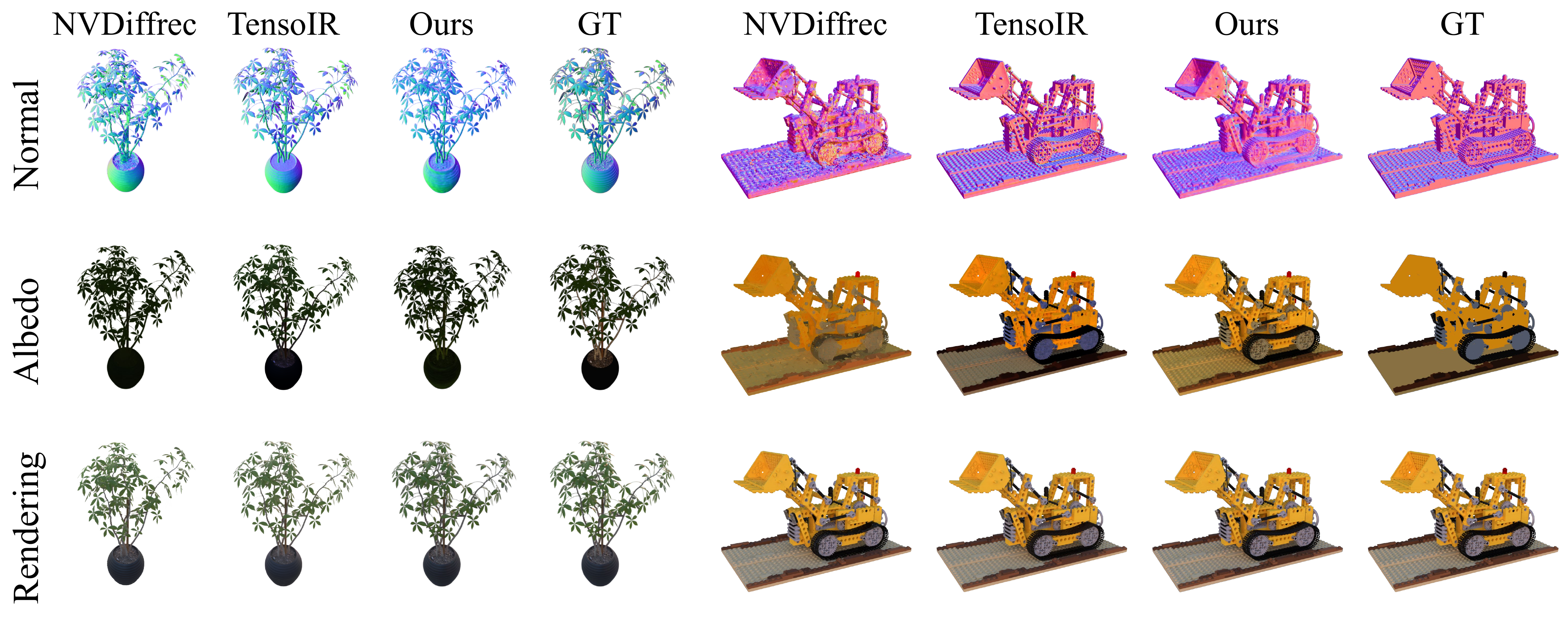}
\caption{\textbf{Qualitative comparison on TensoIR Synthetic}. We visualize the estimated normal, albedo, and rendering results of our GS-IR and baseline methods on two scenes. By utilizing the efficient 3D Gaussian representation and a robust tile-based rasterizer, GS-IR achieves rapid convergence and supports real-time rendering. This performance advantage underscores the effectiveness of our method in addressing complex inverse rendering tasks, thereby surpassing existing state-of-the-art approaches. (For albedo reconstruction results, we follow NeRFactor \cite{zhang2021nerfactor} and scale each RGB channel by a global scalar.)}
\label{fig:comparison}
\end{figure*}

\begin{equation}
\footnotesize
\begin{aligned}
L_\text{s} =& \int_\Omega
\frac{
    DFG
}{
    4 \left<\bm{n} \cdot \bm{l}\right> \left<\bm{n} \cdot \bm{v}\right>
}
L_i(\bm{l}) \left<\bm{l} \cdot \bm{n}\right> d\bm{l} \\
\approx &
\underbrace{
    \int_\Omega \frac{
    DFG
    }{
        4 \left<\bm{n} \cdot \bm{l}\right> \left<\bm{n} \cdot \bm{v}\right>
    } \left<\bm{l} \cdot \bm{n}\right> d\bm{l}
}_{\text{\tiny Environment BRDF -} R}
\underbrace{
    \int_\Omega D\ L_i(\bm{l}) \left<\bm{l} \cdot \bm{n}\right> d\bm{l}
}_{\text{\tiny Pre-Filtered Environment Map - }I_\text{s}},
\end{aligned}
\label{eq:specular}
\end{equation}
where both $R$ and $I_\text{s}$ can be precomputed in advance and stored in look-up tables.
With \cref{eq:diffuse} and \cref{eq:specular}, the rendering results of \cref{eq:shading} can be represented as:
\begin{equation}
\footnotesize
\begin{aligned}
L_o(\bm{x}, \bm{v})\!\! =& L_\text{d} + L_\text{s} \\
\approx& (1 - m)\frac{\bm{a}}{\pi} \left[\left(1 - \text{O}(\bm{x})\right) I^\text{\tiny dir}_\text{d}(\bm{x})\! +\! \text{O}(\bm{x}) I^\text{\tiny indir}_\text{d}(\bm{x})\right]\!\! +\!\!R I_\text{s}.
\end{aligned}
\label{eq:final}
\end{equation}

\begin{table*}[!t]
\centering
\scalebox{0.85}{
\begin{tabular}{c|c|c c c|c c c|c c c|c}
\Xhline{3\arrayrulewidth}
\multirow{2}{*}{Method} &
\multirow{2}{*}{\makecell{Normal \\ MAE} $\downarrow$} &
\multicolumn{3}{c|}{Novel View Synthesis} &
\multicolumn{3}{c|}{Albedo} &
\multicolumn{3}{c|}{Relight} &
\multirow{2}{*}{Runtime} \\
& &
PSNR $\uparrow$ & SSIM $\uparrow$ & LPIPS $\downarrow$ &
PSNR $\uparrow$ & SSIM $\uparrow$ & LPIPS $\downarrow$ &
PSNR $\uparrow$ & SSIM $\uparrow$ & LPIPS $\downarrow$ & \\
\hline
NeRFactor \cite{zhang2021nerfactor} & 6.314 &
24.679 & 0.922 & 0.120 &
25.125 & \cc{3}0.940 & 0.109 &
23.383 & \cc{3}0.908 & 0.131 & $>$ 100 hrs \\
InvRender \cite{zhang2022modeling} & \cc{3}5.074 &
27.367 & 0.934 & 0.089 &
27.341 & 0.933 & \cc{3}0.100 &
\cc{3}23.973 & \cc{2}0.901 & \cc{3}0.101 & 15 hrs \\
NVDiffrec \cite{munkberg2022extracting} & 6.078 &
\cc{3}30.696 & \cc{3}0.962 & \cc{3}0.052 &
\cc{3}29.174 & 0.908 & 0.115 &
19.880 & 0.879 & 0.104 & \cc{1}$<$ 1 hr \\
TensoIR \cite{jin2023tensoir} & \cc{1}4.100 &
\cc{2}35.088 & \cc{1}0.976 & \cc{2}0.040 &
\cc{2}29.275 & \cc{1}0.950 & \cc{2}0.085 &
\cc{1}28.580 & \cc{1}0.944 & \cc{1}0.081 & \cc{3}5 hrs \\
\hline
Ours & \cc{2}4.948 &
\cc{1}35.333 & \cc{2}0.974 & \cc{1}0.039 &
\cc{1}30.286 & \cc{2}0.941 & \cc{1}0.084 &
\cc{2}24.374 & 0.885 & \cc{2}0.096 & \cc{1}$<$ 1 hr \\
\Xhline{3\arrayrulewidth}
\end{tabular}
}
\caption{\textbf{Quantatitive Comparison on TensoIR Synthetic dataset.} Our method outperforms baseline methods in terms of novel view synthesis and albedo quality, showcasing the effectiveness of material decomposition and PBR rendering. This is particularly noteworthy considering that our normal reconstruction is slightly inferior to TensoIR. In terms of relighting performance, we rank second, trailing only behind TensoIR. Importantly, the average training time of our GS-IR is accelerated by a factor of 5x, making its performance acceptable and further demonstrating the effectiveness of our approach in handling complex inverse rendering tasks.}
\label{tab:compairson}
\vspace{-0.2cm}
\end{table*}

For intrinsic decomposition, we optimize the material $\hat{\bm{M}}$ (\ie albedo $\bm{a}$, metallic value $m$, and roughness $\rho$) stored in 3D Gaussians $\mathcal{G}$, environment map $\hat{\bm{E}}$, and illumination volumes $\mathcal{V}^\text{illu}$ by minimizing the decomposition loss $\mathcal{L}_d$:
\begin{equation}
\footnotesize
\begin{aligned}
\mathcal{L}_d =& \underbrace{\left\|\bm{I}- \hat{\bm{I}}^\text{shade}(\hat{\bm{M}}, \hat{\bm{E}}, \mathcal{V}^\text{illu})\right\|}_{\mathcal{L}_\text{shade}} + \underbrace{\lambda_{\bm{M}}\ \mathit{TV}_\text{mat}}_{\mathcal{L}_\text{material}} + \underbrace{\lambda_{\bm{E}}\ \mathit{TV}_\text{light}}_{\mathcal{L}_\text{light}},
\end{aligned}
\label{eq:decompose_loss}
\end{equation}
where $\mathcal{L}_\text{shade}$ indicate the shade loss. $\hat{\bm{I}}^\text{shade}(\hat{\bm{M}}, \hat{\bm{E}}, \mathcal{V}^\text{illu})$ is the recovered image that uses the PBR pipeline defined by \cref{eq:final}.% and then accumulates via volume rendering based on \cref{eq:volrend}.
Please refer to the \emph{supplment} for more details about the material TV loss $\mathcal{L}_\text{material}$ and lighting TV loss  $\mathcal{L}_\text{light}$.
% $\mathcal{L}_\text{material}$ and $\mathcal{L}_\text{light}$ refer to material TV loss and lighting TV loss, respectively, where the definition of $\mathit{TV}(\cdot)$ can be found in \cref{eq:normal_tv_loss}.

%-------------------------------------------------------------------------
\section{Experiments}
\label{sec:experiments}
\noindent\textbf{Dataset \& Metrics} We conduct expermients using benchmark datasets of TensoIR Synthetic \cite{jin2023tensoir} and Mip-NeRF 360 \cite{barron2022mip} for decompositing both objects and scenes. They contain 4 objects with reference materials and $7$ publicly available scenes, respectively. To verify the efficacy of our normal reconstruction, we evaluate the normal quality on the TensoIR Synthetic \cite{jin2023tensoir} dataset using mean angular error (MAE). We further assess our reconstructed albedo quality on this synthetic dataset. More generally, we evaluate the synthesized novel view on both datasets in terms of Peak Signal-to-Noise Ratio (PSNR), Structural Similarity Index Measure (SSIM), and Learned Perceptual Image Patch Similarity (LPIPS) \cite{zhang2018unreasonable}. Note that albedo quality assessment uses the same metrics as novel view synthesis.

\begin{figure}%[htbp]
\centering
\includegraphics[width=0.9\linewidth]{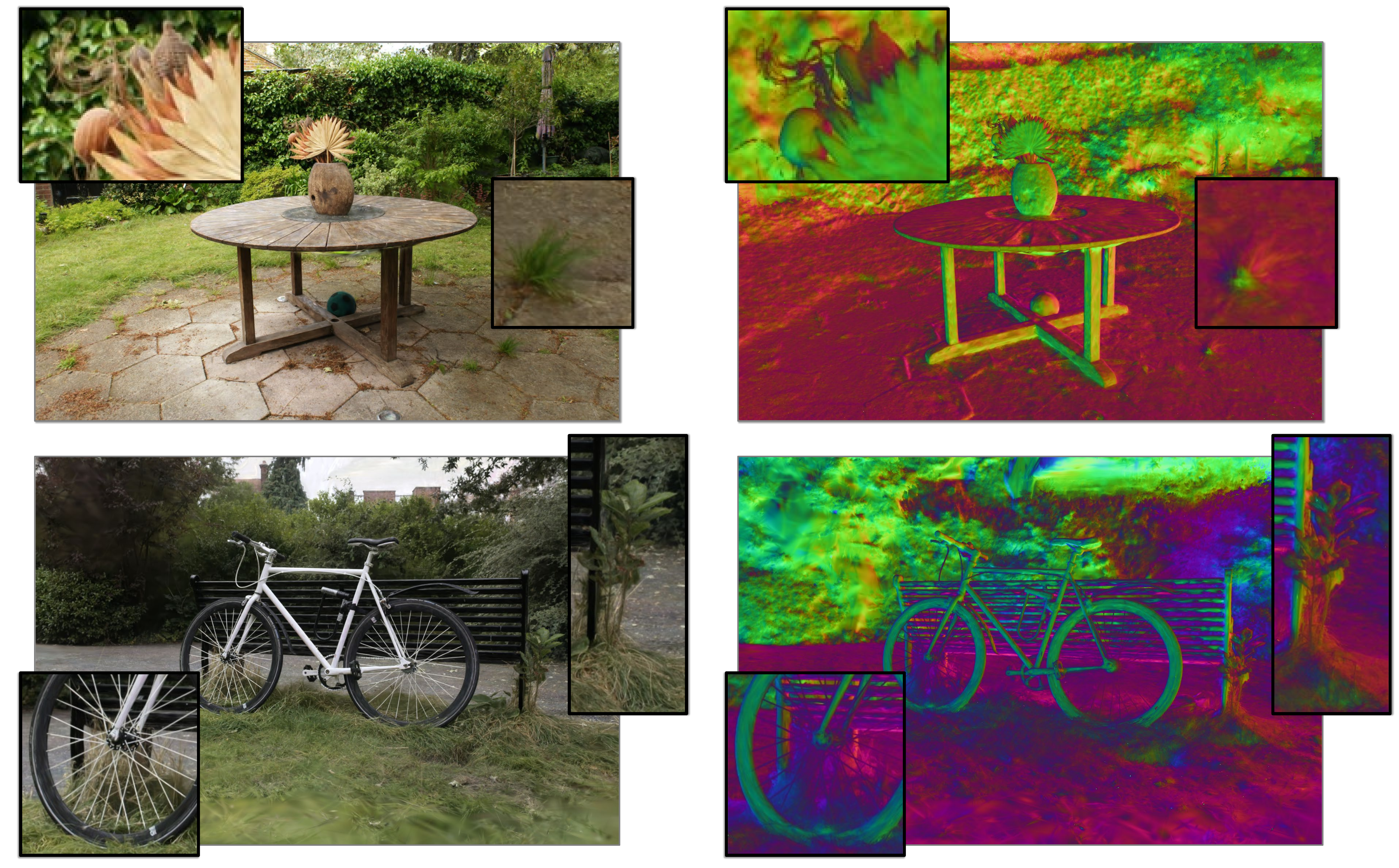}
\caption{\textbf{Novel view synthesis results on Mip-NeRF 360.} GS-IR can reconstruct scene details including geometric normals and high-frequency appearance, rendering high-fidelity appearance and recovering fine geometric details such as those on leaves and bicycle axles. \textbf{Better viewed on screen with zoom in}.}
\label{fig:real}
\vspace{-0.2cm}
\end{figure}

\begin{table}[]
\centering
% \small
\scalebox{0.78}
{
\begin{tabular}{p{0.4\linewidth }|c c c c}
\Xhline{3\arrayrulewidth}
Method & PSNR $\uparrow$ & SSIM $\uparrow$ & LPIPS $\downarrow$ & Runtime $\downarrow$ \\
\hline
NeRF++ \cite{zhang2020nerf++}
& 25.112 & 0.696 & 0.375 & $\approx$ 20h \\
Plenoxels \cite{fridovich2022plenoxels} &
23.079 & 0.625 & 0.462 & $\approx$ 30m \\
INGP-Base \cite{muller2022instant}      &
25.303 & 0.671 & 0.371 & $\approx$ \textbf{5m}\\
INGP-Big \cite{muller2022instant}       &
25.587 & 0.699 & 0.331 & $\approx$ 8m \\
Mip-NeRF 360 \cite{muller2022instant}   &
\cc{1}27.569 & \cc{2}0.793 & \cc{2}0.234 & $\approx$ 48h \\
3DGS \cite{kerbl20233d}                 &
\cc{2}27.21 & \cc{1}0.815 & \cc{1}0.214 & $\approx$ 35m \\
\hline
Ours                                    &
\cc{3}25.381 & \cc{3}0.757 & \cc{3}0.267 & $\approx$ \textbf{45m} \\
\Xhline{3\arrayrulewidth}
\end{tabular}
}
\caption{\textbf{Quantatitive Comparison on Mip-NeRF 360.} The results show that our inverse rendering approach even surpasses some NeRF variants dedicated to novel view synthesis.}
\label{tab:real_comp}
\vspace{-0.2cm}
\end{table}

\subsection{Comparisons}
\label{subsec:comparisons}
We conduct a comprehensive comparison against state-of-the-art neural field-based inverse rendering methods on the public TensoIR Synthetic dataset \cite{jin2023tensoir}. All the methods utilize multi-view images captured under unknown lighting conditions. Our evaluation encompasses normal quality (measured by MAE), novel view synthesis, albedo fidelity, relighting effects (measured by PSNR, SSIM, and LPIPS), and efficiency. \cref{tab:compairson} summarizes the quantitative comparisons on the synthesis dataset.
Our method achieves superior performance in novel view synthesis and albedo quality compared to the baseline methods, demonstrating the effectiveness of material decomposition and PBR rendering, particularly given that our normal reconstruction is slightly inferior to TensoIR. Our relighting performance ranks second, only behind TensoIR. Notably, the average training time of our GS-IR is accelerated by a factor of 5x, making its performance acceptable and demonstrating the effectiveness of our approach in handling complex inverse rendering tasks.
We also include qualitative comparisons in \cref{fig:comparison}, which show that our GS-IR produces reasonable albedo and photorealistic renderings that are closer to the ground truth than most methods. 

Meanwhile, owing to our more efficient and compact representation with powerful expressiveness, our method showcases remarkable performance on complex real unbounded scenes \cite{barron2022mip}. \cref{tab:real_comp} presents the quantitative comparisons on the real dataset. \cref{fig:real} demonstrates the normal reconstruction and novel view synthesis on the real dataset. Our method renders a high-fidelity appearance and recovers fine geometric details, such as those on the leaves and bicycle axil.
In summary, by leveraging the efficient 3D Gaussian representation and a powerful tile-based rasterizer, GS-IR achieves fast convergence and supports real-time rendering. This performance advantage highlights the effectiveness of our method in handling complex inverse rendering tasks, outperforming existing state-of-the-art approaches.

\begin{table}[]
\centering
\scalebox{0.6}{
\begin{tabular}{c|c c c c|c c c}
\Xhline{3\arrayrulewidth}
    \multirow{2}{*}{Method} & \multicolumn{4}{c|}{TensoIR Synthetic \cite{jin2023tensoir}} & \multicolumn{3}{c}{Mip-NeRF 360 \cite{barron2022mip}} \\
    & \makecell{Normal \\ MAE} $\downarrow$ & PSNR $\uparrow$ & SSIM $\uparrow$ & LPIPS $\downarrow$ & PSNR $\uparrow$ & SSIM $\uparrow$ & LPIPS $\downarrow$ \\
    \hline
    % NOTE: this version does not consider `flower` and `treehill` in Mip-NeRF 360
    % Vol. Accum.           & 16.347 & 25.756 & 0.855 & 0.131 & 23.433 & 0.668 & 0.354 \\
    % Peak Selec.           & 9.466  & 28.750 & 0.927 & 0.084 & 24.371 & 0.780 & 0.280 \\
    % Linear Interp.        & 6.218  & 28.983 & 0.939 & 0.066 & 24.487 & 0.759 & 0.282 \\
    % \hline
    % Vol. Accum.$^\dag$    & 9.315  & 30.091 & 0.940 & 0.071 & 25.951 & 0.806 & 0.241 \\
    % Peak Selec.$^\dag$    & 7.986  & 31.064 & 0.950 & 0.060 & 26.431 & 0.812 & 0.243 \\
    % Linear Interp.$^\dag$ & \textbf{4.948}  & \textbf{35.333} & \textbf{0.974} & \textbf{0.039} & \textbf{26.659} & \textbf{0.815} & \textbf{0.229} \\
    Vol. Accum.           & 16.347 & 25.756 & 0.855 & 0.131 & 22.052 & 0.610 & 0.394 \\
    Peak Selec.           & 9.466  & 28.750 & 0.927 & 0.084 & 23.093 & 0.719 & 0.317 \\
    Linear Interp.        & 6.218  & 28.983 & 0.939 & 0.066 & 23.119 & 0.707 & 0.319 \\
    \hline
    Vol. Accum.$^\dag$    & 9.315  & 30.091 & 0.940 & 0.071 & 24.664 & 0.747 & 0.277 \\
    Peak Selec.$^\dag$    & 7.986  & 31.064 & 0.950 & 0.060 & 25.143 & 0.753 & 0.281 \\
    Linear Interp.$^\dag$ & \textbf{4.948}  & \textbf{35.333} & \textbf{0.974} & \textbf{0.039} & \textbf{25.381} & \textbf{0.757} & \textbf{0.267} \\
\Xhline{3\arrayrulewidth}
\end{tabular}
}
\vspace{-0.2cm}
\caption{\textbf{Analyses on the impact of different depth generation strategies on normals.} Methods without $\dag$ marks directly use the normals derived from the depth map; Methods marked with $\dag$ use depth derivation to optimize the normals stored in 3D Gaussians.}
\vspace{-0.2cm}
\label{tab:depth_ablation}
\end{table}

\begin{figure}%[htbp]
\centering
\includegraphics[width=0.9\linewidth]{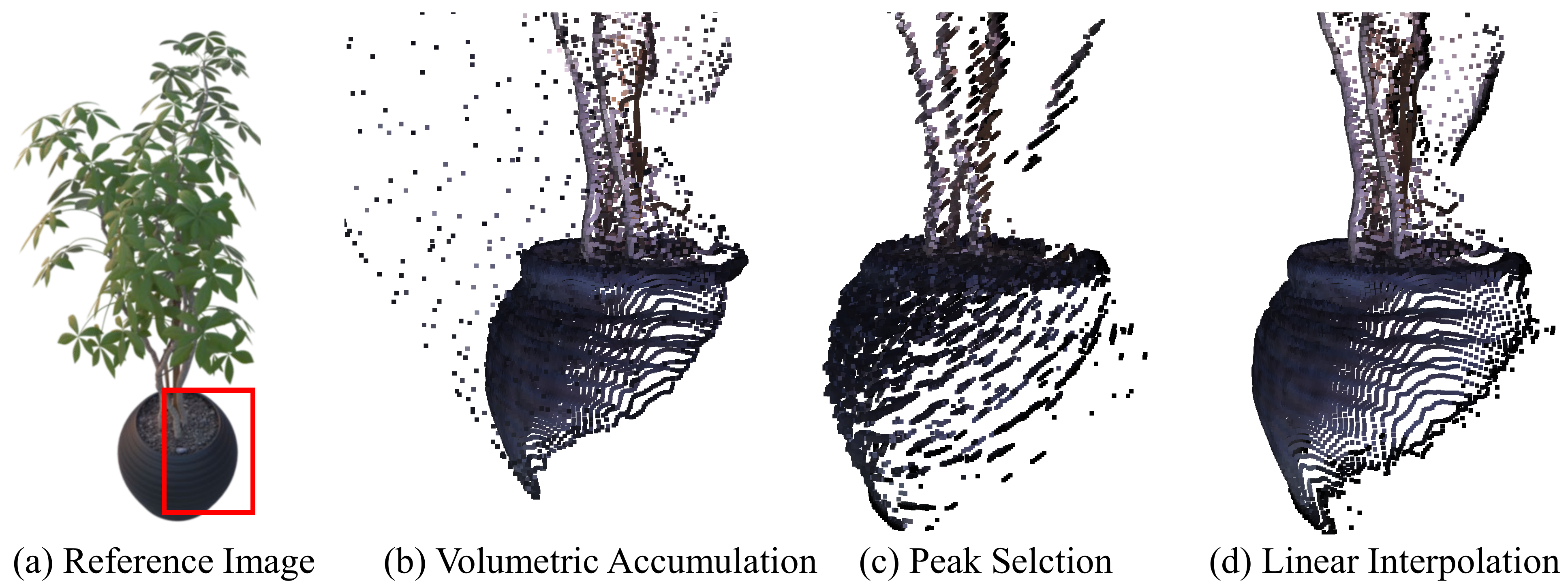}
\vspace{-0.2cm}
\caption{\textbf{Visual comparison of depth produced by different strategies.} The linear interpolation adopted in GS-IR overcomes the \textit{floating} problem and \textit{disc aliasing}.}
\label{fig:depth_ablation}
\vspace{-0.4cm}
\end{figure}

\begin{table}[]
\centering
\scalebox{0.65}{
\begin{tabular}{c|c c c|c c c}
\Xhline{3\arrayrulewidth}
    \multirow{2}{*}{Method} & \multicolumn{3}{c|}{TensoIR Synthetic \cite{jin2023tensoir}} & \multicolumn{3}{c}{Mip-NeRF 360 \cite{barron2022mip}} \\
    & PSNR $\uparrow$ & SSIM $\uparrow$ & LPIPS $\downarrow$ & PSNR $\uparrow$ & SSIM $\uparrow$ & LPIPS $\downarrow$ \\
    \hline
    % NOTE: this version does not consider `flower` and `treehill` in Mip-NeRF 360
    % w/o occlusion       & 34.997 & 0.962 & 0.041 & 25.841 & 0.796 & 0.266 \\
    % w/o indirect illum. & 35.186 & 0.965 & 0.044 & 26.433 & 0.805 & 0.243 \\
    % Ours                & 35.333 & 0.974 & 0.039 & 26.659 & 0.815 & 0.229 \\
    w/o occlusion       & 34.997 & 0.962 & 0.041 & 25.060 & 0.753 & 0.270  \\
    w/o indirect illum. & 35.186 & 0.965 & 0.044 & 24.898 & 0.749 & 0.272 \\
    Ours                & 35.333 & 0.974 & 0.039 & 25.381 & 0.757 & 0.267 \\
\Xhline{3\arrayrulewidth}
\end{tabular}
}
\vspace{-0.2cm}
\caption{\textbf{Analyses on the occlusion and indirect illumination.} Physically modeling indirect illumination improves the inverse rendering of objects and scenes.}
\label{tab:indirect_ablation}
\vspace{-0.2cm}
\end{table}

\begin{figure*}[t]
\centering
\includegraphics[width=0.9\linewidth]{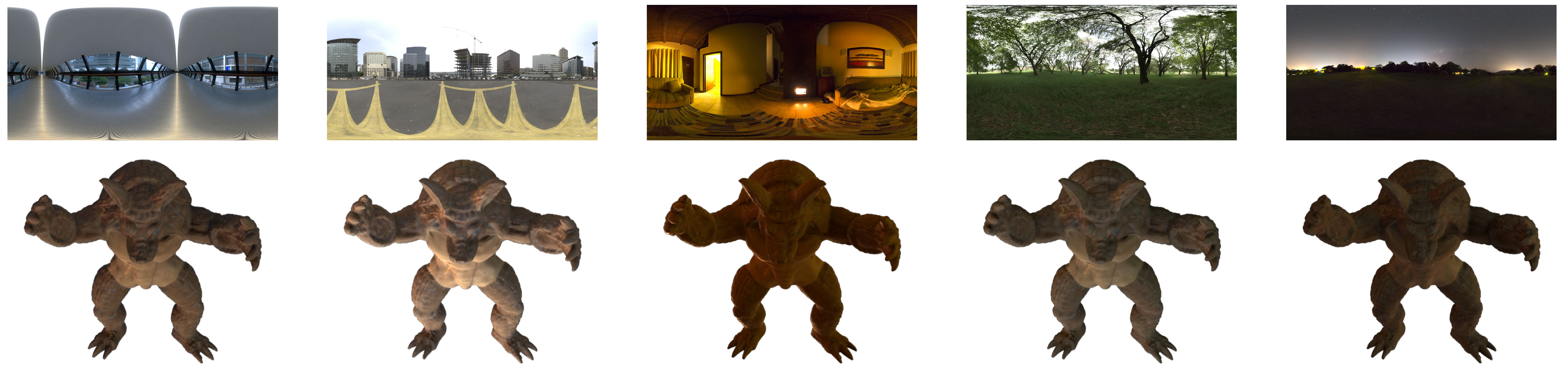}
\vspace{-0.2cm}
\caption{\textbf{Relighting Visualization.} We perform relighting experiments on both synthetic and real scenes using the recovered geometry, material, and illumination properties from our GS-IR method. We test our method under different lighting conditions and directions.}
\label{fig:relighting}
\vspace{-0.2cm}
\end{figure*}

\begin{figure}
\centering
\includegraphics[width=1.0\linewidth]{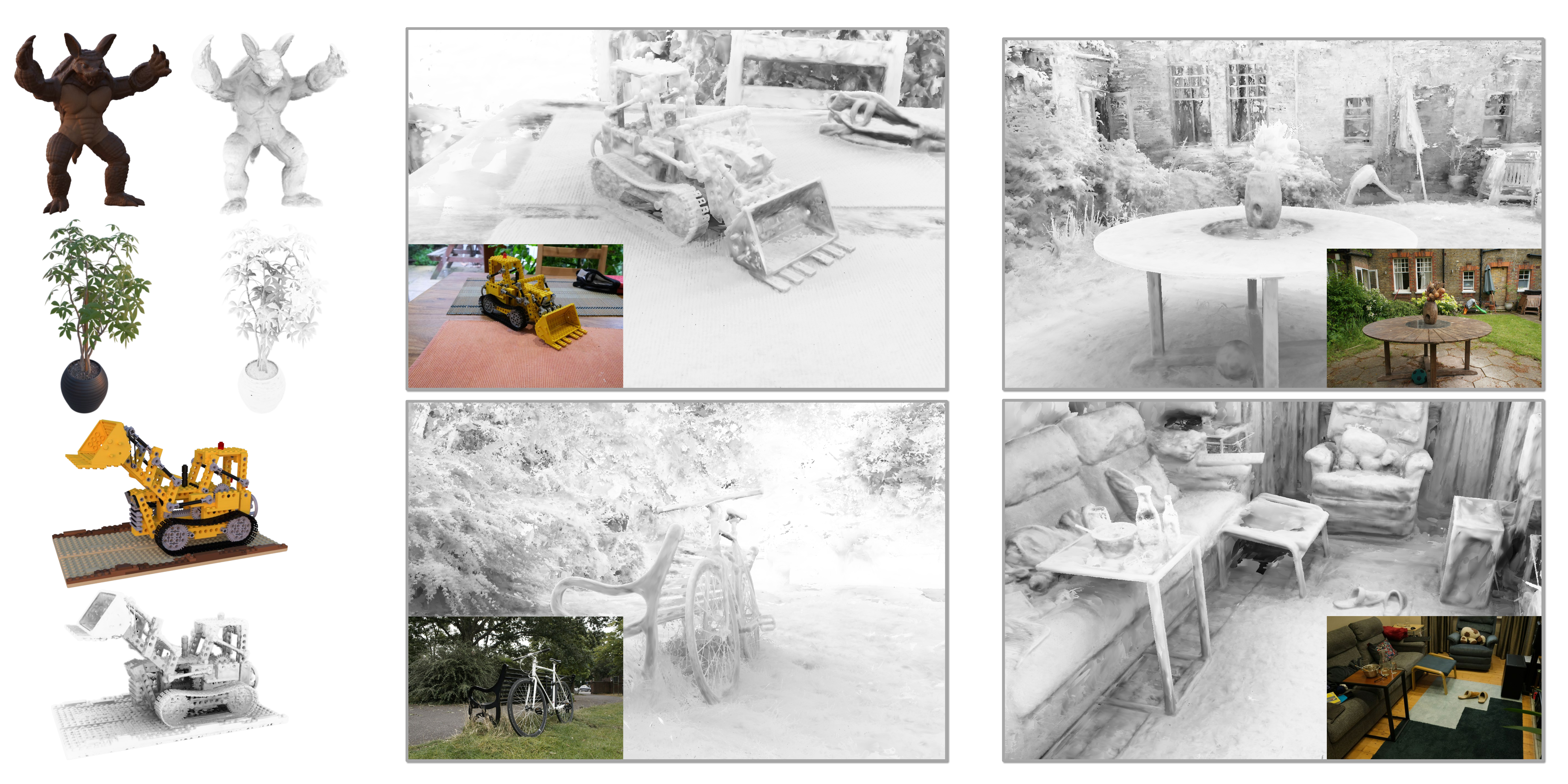}
\vspace{-0.2cm}
\caption{\textbf{Ambient Occlusion Visualization.} The visualization highlights the intricate shadowing and occlusion details captured by our GS-IR method, emphasizing the performance of our approach in modeling indirect illumination.}
\label{fig:occlusion}
\vspace{-0.6cm}
\end{figure}

\subsection{Ablation Studies}
\label{subsec:ablation}
We initially introduce the 3DGS technique for inverse rendering in GS-IR and propose depth-derivation-based normal regularization and a baking-based method to address the challenges encountered during the process. To evaluate the efficacy of our proposed schemes, we design elaborate experiments on both TensoIR Synthetic \cite{jin2023tensoir} and Mip-NeRF 360 \cite{barron2022mip} datasets, providing comprehensive insights into the effectiveness of our approach in handling complex inverse rendering tasks. Below is the detailed ablation study on Normal Regularization and Indirect Illumination.

\noindent\textbf{Analysis on the Normal Regularization} Reliable normal estimation is critical for conducting inverse rendering. To this end, we present depth-derivation-based regularization to facilitate 3D Gaussian-based normal estimation as stated in \cref{subsec:normal}. In this section, we explore the impact of different acquisition schemes on the final normal quality and inverse rendering results. The quantitative results shown in \cref{tab:depth_ablation} demonstrate that using 3D Gaussians as a normal proxy and adopting the linear interpolation strategy significantly improves the normal estimation and inverse rendering results. In addition, \cref{fig:depth_ablation} qualitatively shows that conducting volumetric accumulation results in the \textit{floating} problem (\cf \cref{fig:depth_ablation} (b)). Despite peak selection overcomes this problem, it introduces \textit{disc aliasing} (\cf \cref{fig:depth_ablation} (c)). Compared with them, the linear interpolation adopted in GS-IR robustly produces accurate depth (\cf \cref{fig:depth_ablation} (d)).

\noindent\textbf{Analysis on the Indirect Illumination} 
To demonstrate the effectiveness of our indirect illumination model, we compare our method with two variants: a model without occlusion volume (w/o occlusion) and a model without indirect illumination (w/o indirect illum.). The quantitative comparisons in \cref{tab:indirect_ablation} indicate that each component is crucial for estimating accurate material decomposition and generating photorealistic rendering results. Additionally, \cref{fig:occlusion} showcases the ambient occlusion visualization in both synthetic and real scenes. The visualization highlights the intricate shadowing and occlusion details captured by our GS-IR method, emphasizing the performance of our approach in modeling indirect illumination. This analysis further supports the effectiveness of using occlusion volume and introducing indirect illumination in enhancing the decomposition capabilities of our GS-IR.

\subsection{Application}
\label{subsec:application}
We perform relighting experiments using the recovered geometry, material, and illumination from our GS-IR method. We test GS-IR under different lighting conditions and directions, observing how the reconstructed scene responds to the changes in lighting. The results of these experiments demonstrate that our GS-IR method can effectively handle relighting applications, producing photorealistic renderings under various lighting conditions. More results can be found in the \textit{supplement}.

%------------------------------------------------------------------------
\section{Conclusion}
\label{sec:conclusion}
We present GS-IR, a novel inverse rendering approach based on 3D Gaussian Splatting (3DGS), which employs forward mapping volume rendering to achieve photorealistic novel view synthesis and relighting results. Our GS-IR proposes an optimization scheme with depth-derivation-based regularization for normal estimation and a baking-based occlusion to model indirect lighting. These components are eventually employed to decompose material and illumination. Our extensive experiments demonstrate the effectiveness of GS-IR in achieving state-of-the-art inverse rendering results, surpassing previous neural methods in terms of both reconstruction quality and efficiency.

\noindent\textbf{Limitation} Spherical Harmonics (SH) is only suitable for representing low-frequency, and we only use the occlusion represented by SH to model the diffuse term of indirect illumination. Modeling the specular term of indirect illumination remains a limitation of GS-IR, and has been a challenging problem in computer graphics. We believe it would be valuable to address this limitation in future work and suggest screen space global illumination (SSGI) techniques.

{
    \small
    \bibliographystyle{ieeenat_fullname}
    \bibliography{main}
}

\clearpage
\setcounter{page}{1}
\maketitlesupplementary%%%%%%%%% BODY TEXT
%%%%%%%%% BODY TEXT
\section{Implementation Details}
\label{sec:implementation}
We implement GS-IR in PyTorch framework \cite{paszke2019pytorch} with CUDA extensions, and customized the baking-based method for GS-IR.

\noindent\textbf{Representation.}
In the vanilla GS \cite{kerbl20233d}, each 3D Gaussian utilizes learnable $\mathcal{T} = \{\bm{p}, \bm{s}, \bm{q}\}$ and $\mathcal{A} = \{\alpha, \bm{f}_{c}\}$ to describe its geometric properties and volumetric appearance respectively, where $\bm{p}$ denotes the position vector, $\bm{s}$ denotes the scaling vector, $\bm{q}$ denotes the unit quaternion for rotation, $\alpha$ denotes the opacity and $\bm{f}_{c}$ denotes spherical harmonics (SH) coefficients for view-dependent color.
In GS-IR, we use $\bm{n}$ to present the normal vector of 3D Gaussian and extend the geometric properties as $\mathcal{T} = \{\bm{p}, \bm{s}, \bm{q}, \bm{n}\}$. In addition, we introduce $\mathcal{M} = \{\bm{a}, \rho, m\}$ to describe the material of 3D Gaussian.

\noindent\textbf{Training Details.}
We use the Adam optimizer \cite{kingma2014adam} for training, and the training process includes the initial stage (\cf Sec. 4.1) and decomposition stage (\cf Sec. 4.3).
In the initial stage, we minimize color reconstruction loss $\mathcal{L}_c$ and normal loss $\mathcal{L}_n$ (\cf Eq. (9)) to optimize $\mathcal{T}, \mathcal{A}$ for 30K iterations. In the decomposition stage, we fix $\mathcal{T}, \mathcal{A}$ and minimize the proposed decomposition loss $\mathcal{L}_d$ (\cf  Eq. (16)) to merely optimize $\mathcal{M}$ for 10K iterations. The total optimization is running on a single V100 GPU.

\noindent\textbf{Loss Definition.}
In the initial stage, the supervision loss $\mathcal{L}_\text{init}$ consists of the $L1$ color reconstruction loss $\mathcal{L}_c$ and our proposed normal loss $\mathcal{L}_n$:
\begin{equation}
\begin{aligned}
\mathcal{L}_\text{init} &= \mathcal{L}_c + \mathcal{L}_n \\
\mathcal{L}_n &= \mathcal{L}_{n\text{-}p} + \lambda_{n\text{-}TV} \mathit{TV}_\text{normal}
\end{aligned}
\label{eq:init_loss}
\end{equation}
the smoothing term $\mathit{TV}_\text{normal}$ in our proposed normal loss $\mathcal{L}_n$ is a total variation (TV) loss conditioned by the predicted normal map $\hat{\bm{N}}$ and the given reference image $\bm{I}$:
\begin{equation}\footnotesize
\begin{aligned}
\triangle^{\hat{\bm{N}}}_{ij} = \exp&\left(
-\vert\bm{I}_{i,j} - \bm{I}_{i-1,j}\vert
\right)(\hat{\bm{N}}_{i,j} - \hat{\bm{N}}_{i-1,j})^2\ + \\
\exp&\left(
-\vert\bm{I}_{i,j} - \bm{I}_{i,j-1}\vert
\right)(\hat{\bm{N}}_{i,j} - \hat{\bm{N}}_{i,j-1})^2, \\
\mathit{TV}_\text{normal}
&=\frac{1}{\vert\hat{\bm{N}}\vert}\sum_{i,j} \triangle^{\hat{\bm{N}}}_{ij}.
% \sum_{\bm{u}\in \bm{U}}\triangle^2(\hat{\bm{n}}_{\bm{u}}),
\end{aligned}
\label{eq:normal_tv_loss}
\end{equation}

In the decomposition stage, the supervision loss $\mathcal{L}_d$ includes $\mathcal{L}_\text{shade}, \mathcal{L}_\text{material}, \mathcal{L}_\text{light}$:
\begin{equation}\footnotesize
\begin{aligned}
    \mathcal{L}_d =& \underbrace{\left\|\bm{I}- \hat{\bm{I}}^\text{shade}(\hat{\bm{M}}, \hat{\bm{E}}, \mathcal{V}^\text{illu})\right\|}_{\mathcal{L}_\text{shade}} + \underbrace{\lambda_{\bm{M}}\ \mathit{TV}_\text{mat}}_{\mathcal{L}_\text{material}} + \underbrace{\lambda_{\bm{E}}\ \mathit{TV}_\text{light}}_{\mathcal{L}_\text{light}},
\end{aligned}
\label{eq:supp_decompose_loss}
\end{equation}
the smoothing term $\mathit{TV}_\text{mat}$ in \cref{eq:supp_decompose_loss} is a TV loss similar to $\mathit{TV}_\text{normal}$ in \cref{eq:normal_tv_loss}:
\begin{equation}\footnotesize
\begin{aligned}
\triangle^{\hat{\bm{M}}}_{ij} = \exp&\left(
-\vert\bm{I}_{i,j} - \bm{I}_{i-1,j}\vert
\right)(\hat{\bm{M}}_{i,j} - \hat{\bm{M}}_{i-1,j})^2\ + \\
\exp&\left(
-\vert\bm{I}_{i,j} - \bm{I}_{i,j-1}\vert
\right)(\hat{\bm{M}}_{i,j} - \hat{\bm{M}}_{i,j-1})^2, \\
\mathit{TV}_\text{mat}
&=\frac{1}{\vert\hat{\bm{M}}\vert}\sum_{i,j} \triangle^{\hat{\bm{M}}}_{ij},
\end{aligned}
\label{eq:material_tv_loss}
\end{equation}
where $\hat{\bm{M}}$ is the predicted material map. Unlike the above two smoothing terms, $\mathit{TV}_\text{light}$ is defined as:
\begin{equation}\footnotesize
\begin{aligned}
\triangle^{\hat{\bm{E}}}_{ij} = &(\hat{\bm{E}}_{i,j} - \hat{\bm{E}}_{i-1,j})^2\ + (\hat{\bm{E}}_{i,j} - \hat{\bm{E}}_{i,j-1})^2, \\
\mathit{TV}_\text{light}
&=\frac{1}{\vert\hat{\bm{E}}\vert}\sum_{i,j} \triangle^{\hat{\bm{E}}}_{ij}.
\end{aligned}
\label{eq:light_tv_loss}
\end{equation}

During training, we set $\lambda_{n\text{-}TV}, \lambda_{\bm{M}}, \lambda_{\bm{E}}$ to $5.0, 1.0, 0.01$. And we study the efficacy of these smoothing terms in \cref{sec:ablation_on_loss}.

\begin{figure*}[!tb]
\centering
\begin{minipage}[t]{1.0\linewidth}
    \centering
    \includegraphics[width=\textwidth]{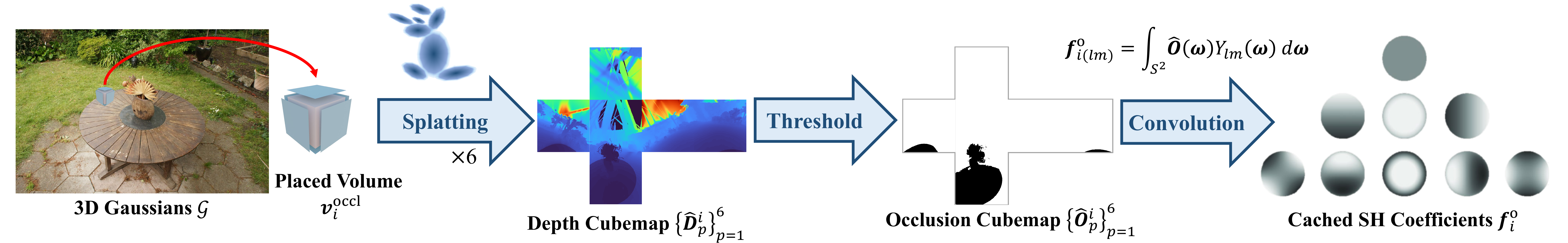}
    \subcaption{Occlusion caching from the pretrained 3D Gaussians $\mathcal{G}$ in the baking stage.}
    \label{fig:caching}
\end{minipage} \quad
\begin{minipage}[t]{1.0\linewidth}
    \includegraphics[width=\textwidth]{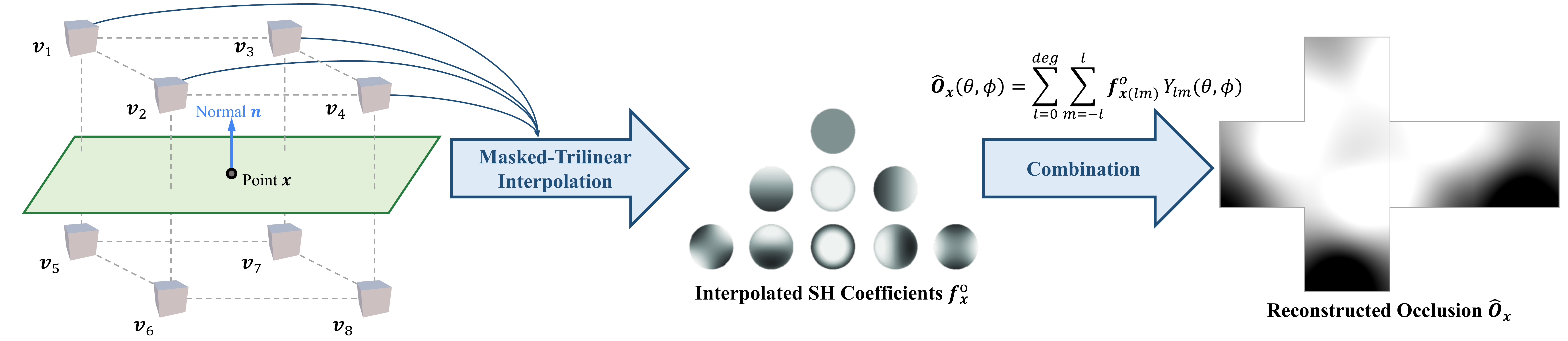}
    \subcaption{Occlusion recovery from occlusion volumes $\mathcal{V}^\text{occl}$ in the decomposition stage.}
    \label{fig:recovery}
\end{minipage}
\caption{Occlusion caching and recovery in GS-IR.
}
\label{fig:occlsion_recovery}
\end{figure*}

\begin{table*}[!tb]
\centering
\scalebox{0.88}{
\begin{tabular}{l c|c|c c c|c c c|c c c}
\Xhline{3\arrayrulewidth}
\multirow{2}{*}{Scene} &
\multirow{2}{*}{Method} &
\multirow{2}{*}{\makecell{Normal \\ MAE} $\downarrow$} &
\multicolumn{3}{c|}{Novel View Synthesis} &
\multicolumn{3}{c|}{Albedo} &
\multicolumn{3}{c}{Relight} \\
& & &
PSNR $\uparrow$ & SSIM $\uparrow$ & LPIPS $\downarrow$ &
PSNR $\uparrow$ & SSIM $\uparrow$ & LPIPS $\downarrow$ &
PSNR $\uparrow$ & SSIM $\uparrow$ & LPIPS $\downarrow$ \\
\hline
\multirow{5}{*}{Lego}
& NeRFactor & 9.767 &
26.076 & 0.881 & 0.151 &
25.444 & 0.937 & 0.112 &
23.246 & 0.865 & 0.156 \\
& InvRender & 9.980 &
24.391 & 0.883 & 0.151 &
21.435 & 0.882 & 0.160 &
20.117 & 0.832 & 0.171 \\
& NVDiffrec & 12.486 &
30.056 & 0.945 & 0.059 &
21.353 & 0.849 & 0.166 &
20.088 & 0.844 & 0.114 \\
& TensoIR   & 5.980  &
34.700 & 0.968 & 0.037 &
25.240 & 0.900 & 0.145 &
28.581 & 0.944 & 0.081  \\
& Ours      & 8.078  &
34.379 & 0.968 & 0.036 &
24.958 & 0.889 & 0.143 &
23.256 & 0.842 & 0.117 \\
\hline
\multirow{5}{*}{Hotdog}
& NeRFactor & 5.579 &
24.498 & 0.940 & 0.141 &
24.654 & 0.950 & 0.142 &
22.713 & 0.914 & 0.159 \\
& InvRender & 3.708 &
31.832 & 0.952 & 0.089 &
27.028 & 0.950 & 0.094 &
27.630 & 0.928 & 0.089 \\
& NVDiffrec & 5.068 &
34.903 & 0.972 & 0.054 &
26.057 & 0.920 & 0.116 &
19.075 & 0.885 & 0.118 \\
& TensoIR   & 4.050 &
36.820 & 0.976 & 0.045 &
30.370 & 0.947 & 0.093 &
27.927 & 0.933 & 0.115 \\
& Ours      & 4.771 &
34.116 & 0.972 & 0.049 &
26.745 & 0.941 & 0.088 &
21.572 & 0.888 & 0.140 \\
\hline
\multirow{5}{*}{Armadillo}
& NeRFactor & 3.467 &
26.479 & 0.947 & 0.095 &
28.001 & 0.946 & 0.096 &
26.887 & 0.944 & 0.102 \\
& InvRender & 1.723 &
31.116 & 0.968 & 0.057 &
35.573 & 0.959 & 0.076 &
27.814 & 0.949 & 0.069  \\
& NVDiffrec & 2.190 &
33.664 & 0.983 & 0.031 &
38.844 & 0.969 & 0.076 &
23.099 & 0.921 & 0.063 \\
& TensoIR   & 1.950 &
39.050 & 0.986 & 0.039 &
34.360 & 0.989 & 0.059 &
34.504 & 0.975 & 0.045 \\
& Ours      & 2.176 &
39.287 & 0.980 & 0.039 &
38.572 & 0.986 & 0.051 &
27.737 & 0.918 & 0.091 \\
\hline
\multirow{5}{*}{Ficus}
& NeRFactor & 6.442 &
21.664 & 0.919 & 0.095 &
22.402 & 0.928 & 0.085 &
20.684 & 0.907 & 0.107 \\
& InvRender & 4.884 &
22.131 & 0.934 & 0.057 &
25.335 & 0.942 & 0.072 &
20.330 & 0.895 & 0.073 \\
& NVDiffrec & 4.567 &
22.131 & 0.946 & 0.064 &
30.443 & 0.894 & 0.101 &
17.260 & 0.865 & 0.073 \\
& TensoIR   & 4.420 &
29.780 & 0.973 & 0.041 &
27.130 & 0.964 & 0.044 &
24.296 & 0.947 & 0.068 \\
& Ours      & 4.762 &
33.551 & 0.976 & 0.031 &
30.867 & 0.948 & 0.053 &
24.932 & 0.893 & 0.081 \\
\hline
\Xhline{3\arrayrulewidth}
\end{tabular}
}
\caption{Per-scene results on TensoIR Synthetic dataset. For albedo reconstruction results, we follow NeRFactor \cite{zhang2021nerfactor} and scale each RGB channel by a global scalar.}
\label{tab:TensoIR}
\end{table*}

\begin{figure*}[!tb]
\centering
\includegraphics[width=\textwidth]{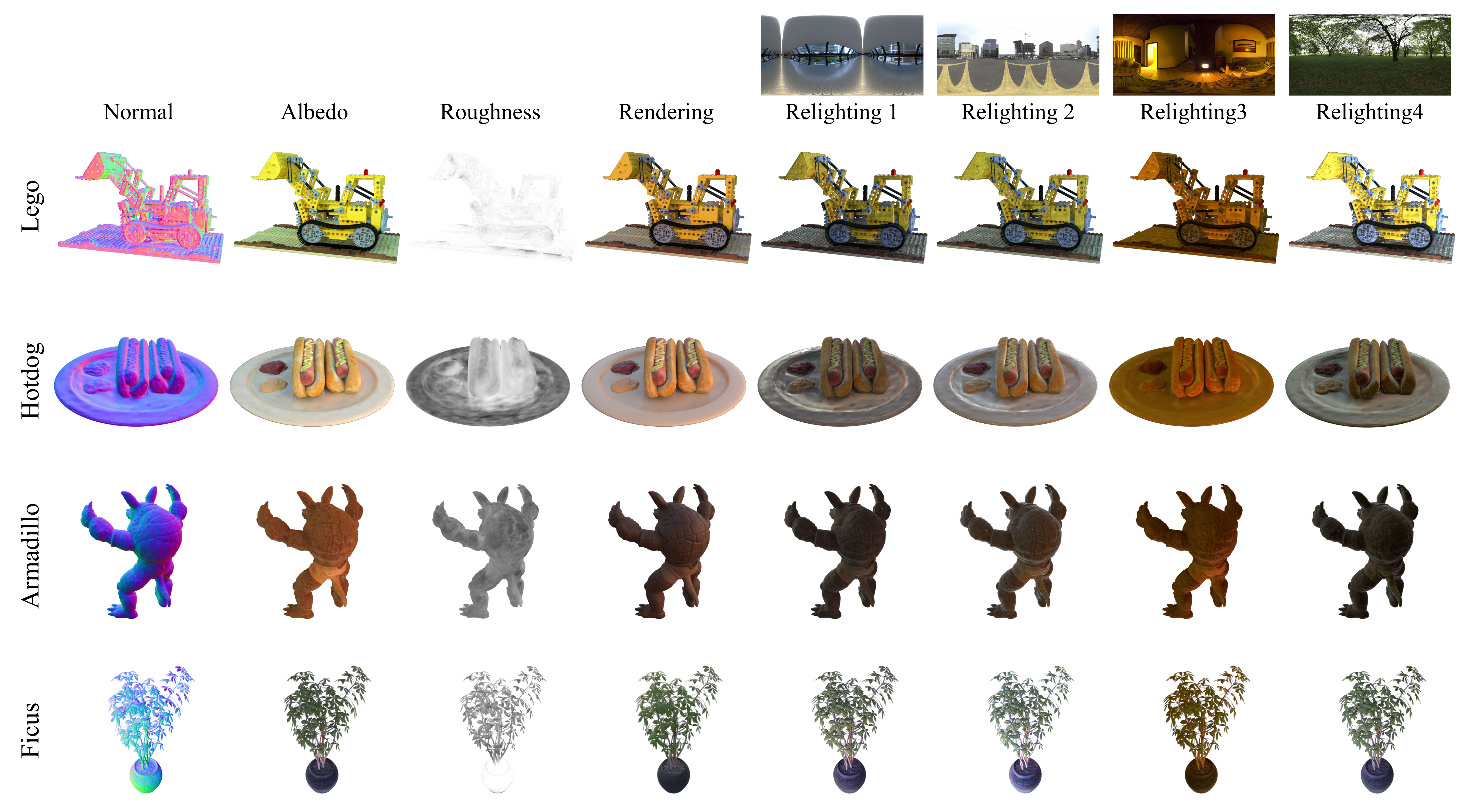}
\caption{Visualization of our inverse rendering and relighting results on TensoIR Synthetic dataset.
}
\label{fig:tensoir}
\end{figure*}

\section{Occlusion Caching and Recovery}
\label{sec:occlusion}
In the baking stage (\cf Sec. 4.2), we introduce SH architectures and cache occlusion into occlusion volumes $\mathcal{V}^\text{occl}$ as illustrated in \cref{fig:caching}. For each volume $\bm{v}^\text{occl}_i \subset \mathcal{V}^\text{occl}$,  we set six cameras with FoV of $90^{\degree}$ and non-overlapping each other, and perform six render passes to obtain the depth cubemap $\{\hat{\bm{D}}^i_p\}^6_{p=1}$. Then we convert $\{\hat{\bm{D}}^i_p\}^6_{p=1}$ into the occlusion cubemap $\{\hat{\bm{O}}^i_p\}^6_{p=1}$ and store the principal components of occlusion into SH coefficients $\bm{f}^o_{\bm{x}}$.

In the decomposition stage, we recover the ambient occlusion (AO) for each surface point $\bm{x}$ from occlusion volumes $\mathcal{V}^\text{occl}$. The first step is to get the coefficients $\bm{f}^o_{\bm{x}}$ of the point $\bm{x}$. Considering that AO of the point $\bm{x}$ only calculates the occlusion integral of the upper hemisphere $\Omega$ of the normal $\bm{n}$, we thus conduct masked-trilinear interpolation to get the correct coefficients.
As illustrated in \cref{fig:recovery}, for the given point $\bm{x}$ with normal $\bm{n}$, we firstly find the eight nearest volumes $\{\bm{v}_{k}\}^8_{k=1}$. In this case, each volume has position vector $\bm{p}_k$ and SH coefficients $\bm{f}^o_{k}$.
Given the trilinear interpolation weights $\left\{w_{k}\right\}^8_{k=1}$ \footnote{The weights in trilinear interpolation satisfy $\sum^8_{k=1} w_k = 1$} defined in vanilla trilinear interpolation, we get the coefficients $\bm{f}^\text{o}_{\bm{x}}$:%the weights $\left\{\hat{w}_k\right\}^8_{k=1}$ in masked-trilinear interpolation are defined as:
\begin{equation}
\small
\begin{aligned}
\tilde{w}_{k} =& 
\begin{cases}
0, & (\bm{p}_k - \bm{x}) \cdot \bm{n} \leq 0 \\
w_k, & (\bm{p}_k - \bm{x}) \cdot \bm{n} > 0
\end{cases}, \\
\hat{w}_k =& \frac{\tilde{w}_k}{\sum^8_{k=1}\tilde{w}_k}, \\
\bm{f}^\text{o}_{\bm{x}(lm)} =& \sum^8_{k=1} \hat{w}_k \bm{f}^\text{o}_{k(lm)}.
\end{aligned}
\vspace{-0.1in}
\end{equation}

After performing masked-trilinear interpolation, %we get the SH coefficient $\bm{f}^o_{\bm{x}}$ of the point $\bm{x}$,
the occlusion $\hat{\bm{O}}_{\bm{x}}$ is written as:
\begin{equation}\small
\hat{\bm{O}}_{\bm{x}}(\theta, \phi) = \sum^{deg}_{l=0}\sum^l_{m=-l} \bm{f}^o_{\bm{x}(lm)} Y_{lm}(\theta, \phi).
\vspace{-0.1in}
\end{equation}

For indirect illumination $I^\text{indir}_\text{d}$ in Eq. (13), we recover it from the volumes $\mathcal{V}^\text{illu}$ via vanilla trilinear interpolation.

\begin{table}[htb]
\renewcommand\arraystretch{1.3}
\centering
\scalebox{0.53}{
\begin{tabular}{l|c c c c c|c c c c}
\Xhline{3\arrayrulewidth}
Method & bicycle & flowers & garden & stump & treehill &
room & counter & kitchen & bonsai \\
\hline
NeRF++
& 22.64 & 20.31 & 24.32 & 24.34 & 22.20
& 28.87 & 26.38 & 27.80 & 29.15 \\
Plenoxels
& 21.91 & 20.10 & 23.49 & 20.66 & 22.25
& 27.59 & 23.62 & 23.42 & 24.67 \\
INGP-Base
& 22.19 & 20.35 & 24.60 & 23.63 & 22.36
& 29.27 & 26.44 & 28.55 & 30.34 \\
INGP-Big
& 22.17 & 20.65 & 25.07 & 23.47 & 22.37
& 29.69 & 26.69 & 29.48 & 30.69 \\
Mip-NeRF 360
& 24.40 & 21.64 & 26.94 & 26.36 & 22.81
& 29.69 & 26.69 & 29.48 & 30.69 \\
3DGS
& 25.25 & 21.52 & 27.41 & 26.55 & 22.49
& 30.63 & 28.70 & 30.32 & 31.98 \\
\hline
Ours
& 23.80 & 20.57 & 25.72 & 25.37 & 21.79
& 28.79 & 26.22 & 27.99 & 28.18 \\
\Xhline{3\arrayrulewidth}
\end{tabular}
}
\caption{PSNR scores for Mip-NeRF360 scenes.}
\vspace{-0.6cm}
\label{tab:360psnr}
\end{table}

\begin{table}[htb]
\renewcommand\arraystretch{1.3}
\centering
\scalebox{0.53}{
\begin{tabular}{l|c c c c c|c c c c}
\Xhline{3\arrayrulewidth}
Method & bicycle & flowers & garden & stump & treehill &
room & counter & kitchen & bonsai \\
\hline
NeRF++
& 0.526 & 0.453 & 0.635 & 0.594 & 0.530
& 0.530 & 0.802 & 0.816 & 0.876 \\
Plenoxels
& 0.496 & 0.431 & 0.606 & 0.523 & 0.509
& 0.842 & 0.759 & 0.648 & 0.814 \\
INGP-Base
& 0.491 & 0.450 & 0.649 & 0.574 & 0.518
& 0.855 & 0.798 & 0.818 & 0.890 \\
INGP-Big
& 0.512 & 0.486 & 0.701 & 0.594 & 0.542
& 0.871 & 0.817 & 0.858 & 0.906 \\
Mip-NeRF 360
& 0.693 & 0.583 & 0.816 & 0.746 & 0.632
& 0.913 & 0.895 & 0.920 & 0.939 \\
3DGS
& 0.771 & 0.605 & 0.868 & 0.775 & 0.638
& 0.914 & 0.905 & 0.922 & 0.938 \\
\hline
Ours
& 0.706 & 0.543 & 0.804 & 0.716 & 0.586
& 0.867 & 0.839 & 0.867 & 0.883 \\
\Xhline{3\arrayrulewidth}
\end{tabular}
}
\caption{SSIM scores for Mip-NeRF360 scenes.}
\vspace{-0.6cm}
\label{tab:360ssim}
\end{table}

\begin{table}[htb]
\renewcommand\arraystretch{1.3}
\centering
\scalebox{0.53}{
\begin{tabular}{l|c c c c c|c c c c}
\Xhline{3\arrayrulewidth}
Method & bicycle & flowers & garden & stump & treehill &
room & counter & kitchen & bonsai \\
\hline
NeRF++
& 0.455 & 0.466 & 0.331 & 0.416 & 0.466
& 0.335 & 0.351 & 0.260 & 0.291 \\
Plenoxels
& 0.506 & 0.521 & 0.386 & 0.503 & 0.540 
& 0.419 & 0.441 & 0.447 & 0.398 \\
INGP-Base
& 0.487 & 0.481 & 0.312 & 0.450 & 0.489
& 0.301 & 0.342 & 0.254 & 0.227 \\
INGP-Big
& 0.446 & 0.441 & 0.257 & 0.421 & 0.450
& 0.261 & 0.306 & 0.195 & 0.205 \\
Mip-NeRF 360
& 0.289 & 0.345 & 0.164 & 0.254 & 0.338
& 0.211 & 0.203 & 0.126 & 0.177 \\
3DGS
& 0.205 & 0.336 & 0.103 & 0.210 & 0.317
& 0.220 & 0.204 & 0.129 & 0.205 \\
\hline
Ours
& 0.259 & 0.371 & 0.158 & 0.258 & 0.372
& 0.279 & 0.260 & 0.188 & 0.264 \\
\Xhline{3\arrayrulewidth}
\end{tabular}
}
\caption{LPIPS scores for Mip-NeRF360 scenes.}
\vspace{-0.6cm}
\label{tab:360lpips}
\end{table}

\section{Results on TensoIR Synthetic Dataset}
\label{sec:results_tensoir}
\cref{tab:TensoIR} provides the results on normal estimation, novel view synthesis, albedo reconstruction, and relighting for all four scenes. We also visualize the inverse rendering and relighting results of GS-IR in \cref{fig:tensoir}.

\section{Results on Mip-NeRF 360}
\label{sec:results_mipnerf360}
\begin{figure*}[!t]
\centering
\includegraphics[width=\textwidth]{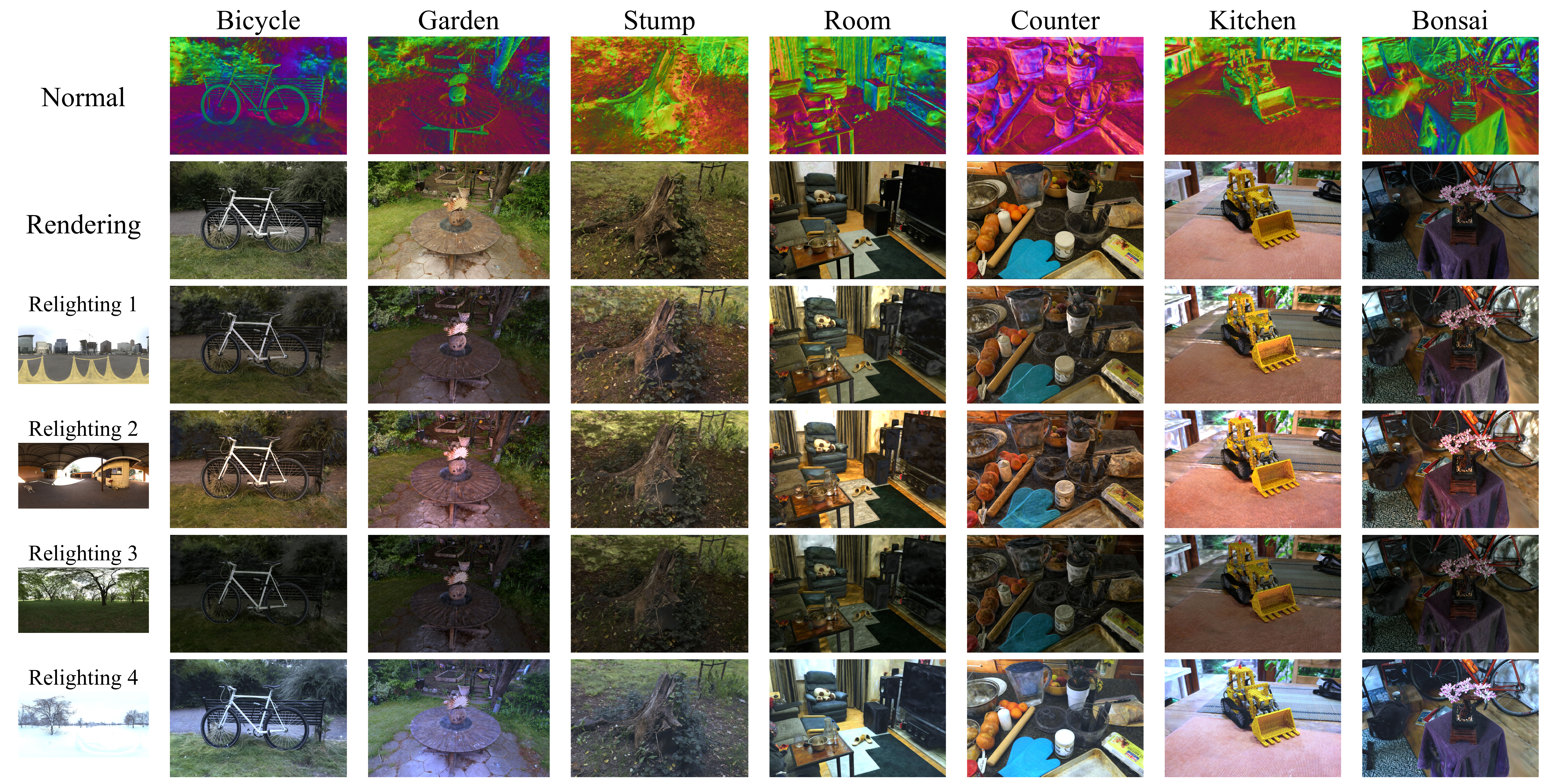}
\caption{Visualization of our inverse rendering and relighting results on the Mip-NeRF 360 dataset.
}
\label{fig:mipnerf360}
\end{figure*}

For Mip-NeRF 360 \cite{barron2022mip}, a dataset captured from the real world, we list the results on novel view synthesis (\ie PSNR, SSIM, and LPIPS) of GS-IR and some NeRF variants \cite{zhang2020nerf++, fridovich2022plenoxels, muller2022instant} in \cref{tab:360psnr,tab:360ssim,tab:360lpips}. In addition, we provide the normal estimation, novel view synthesis, and relighting results of all seven publicly available scenes in \cref{fig:mipnerf360}.

\section{Ablation on Loss}
\label{sec:ablation_on_loss}
The loss in GS-IR consists of contrast terms and smoothing terms. For contrast terms, we set the weights of color reconstruction loss $\mathcal{L}_c$, normal penalty loss $\mathcal{L}_{n\text{-}p}$, and shade loss $\mathcal{L}_\text{shade}$ to 1, which is intuitive.
And the smoothing terms include $\mathit{TV}_\text{normal}$, $\mathit{TV}_\text{mat}$, and $\mathit{TV}_\text{light}$,
we evaluate their efficacy by adjusting their weights (\ie $\lambda_{n\text{-}TV}$, $\lambda_{\bm{E}}$, and $\lambda_{\bm{M}}$), and the ablation results are shown in \cref{tab:loss_ablation}.

\begin{table}[htb]
\centering
\scalebox{0.58}{
\begin{tabular}{c c c|c|c c c|c c c}
\Xhline{3\arrayrulewidth}
& & &
Normal &
\multicolumn{3}{c|}{Novel View Synthesis} &
\multicolumn{3}{c}{Albedo} \\
$\lambda_{n\text{-}TV}$ & $\lambda_{\bm{E}}$ & $\lambda_{\bm{M}}$ &
MAE $\downarrow$ &
PSNR $\uparrow$ & SSIM $\uparrow$ & LPIPS $\downarrow$ &
PSNR $\uparrow$ & SSIM $\uparrow$ & LPIPS $\downarrow$ \\
\hline
& & & 5.030 &
35.170 & 0.970 & 0.042 &
30.083 & 0.938 & 0.090 \\
\checkmark & & & \textbf{4.948} &
35.330 & \textbf{0.974} & 0.039 &
30.216 & 0.940 & 0.088 \\
\checkmark & \checkmark & & \textbf{4.948} &
35.230 & 0.972 & 0.040 &
30.236 & 0.940 & 0.087 \\
\checkmark & & \checkmark & \textbf{4.948} &
35.314 & 0.973 & \textbf{0.038} &
30.275 & \textbf{0.941} & 0.085 \\
\checkmark & \checkmark & \checkmark & \textbf{4.948} &
\textbf{35.333} & \textbf{0.974} & 0.039 &
\textbf{30.286} & \textbf{0.941} & \textbf{0.084} \\
\Xhline{3\arrayrulewidth}
\end{tabular}
}
\caption{\textbf{Analysis of the impact of different loss terms on the TensoIR dataset.} \checkmark indicates setting the smoothing term to be valid.}
\label{tab:loss_ablation}
\end{table}


\end{document}